\PassOptionsToPackage{unicode}{hyperref}
\PassOptionsToPackage{naturalnames}{hyperref}

\documentclass[12pt,english]{article}
\usepackage{mathptmx}
\usepackage[T1]{fontenc}
\usepackage{geometry}
\usepackage{color}
\usepackage{babel}
\usepackage{booktabs}
\usepackage{mathtools}
\usepackage{amsmath}
\usepackage{stmaryrd}
\usepackage{graphicx}
\usepackage{setspace}
\usepackage{esint}
\usepackage{newtxtext,newtxmath}
\usepackage[authormarkup=none,final]{changes}
\definechangesauthor[name={del}, color=red]{del}
\usepackage[unicode=true,pdfusetitle,
bookmarks=true,bookmarksnumbered=false,bookmarksopen=false,breaklinks=false,pdfborder={0 0 1},backref=false,colorlinks=true]
 {hyperref}
\hypersetup{
 linkcolor=blue,urlcolor={blue},filecolor={blue},citecolor={blue}}

\makeatletter

\newcommand{\lyxaddress}[1]{
	\par {\raggedright #1
	\noindent\par}
}

\usepackage{caption}
\usepackage{subcaption}
\usepackage{cleveref}
\crefformat{section}{\S#2#1#3}

\usepackage{chngcntr}
\usepackage{mathtools}
\counterwithout{figure}{section}
\counterwithout{equation}{section}
\usepackage{times}
\usepackage{amsmath}
\usepackage{floatrow}	
\usepackage{times}
\usepackage{upgreek}
\usepackage{hyperref} 
\usepackage{floatrow}
\counterwithout{figure}{section}
\counterwithout{equation}{section}
\usepackage{changepage}
\usepackage{babel}
\counterwithout{figure}{section}
\counterwithout{equation}{section}
\usepackage{times}
\usepackage{float}
\usepackage{bm}
\usepackage[numbers,sort&compress]{natbib} 
\date{}
\renewcommand{\fnum@figure}{Fig. \thefigure}
\makeatother
\begin{document}

\global\long\def\kI{k^{I}}
\global\long\def\kII{k^{II}}
\global\long\def\ks{k^{s}}

\title{Dynamic single-input control of multi-state multi-transition soft robotic actuator}
\author{Geron Yamit$^*$, Ben-Haim Eran$^*$, Gat D.~Amir, Or Yizhar, Givli Sefi$^\dag$}

\maketitle
\def\thefootnote{*}
\footnotetext{These authors contributed equally to this work}

\lyxaddress{\begin{center}
Faculty of Mechanical Engineering, Technion --Israel Institute of
Technology, Haifa 32000, Israel\\
$\dag$ Corresponding author: givli@technion.ac.il
\par\end{center}}

\begin{abstract}
Soft robotics is an attractive and rapidly emerging field, in which actuation is coupled with the elastic response of the robot’s structure to achieve complex deformation patterns. A crucial challenge is the need for multiple control inputs, which adds significant complication to the system. We propose a novel concept of single-input control of an actuator composed of interconnected bi-stable elements. Dynamic response of the actuator and pre-designed differences between the elements are exploited to facilitate any desired multi-state transition, using a single dynamic input. We show formulation and analysis of the control system’s dynamics and pre-design of its multiple equilibrium states, as well as their stability. Then we fabricate and demonstrate experimentally on single-input control of two- and four-element actuators, where the latter can achieve transitions between up to 48 desired states. Our work paves the way for next-generation soft robotic actuators with minimal actuation and maximal dexterity.
\end{abstract}

\section{Introduction}
\label{sec:Introduction}

Robots that carry out complicated tasks, such as dexterous robots for medical procedures, articulated robotic arms for manipulation and grasping tasks, or snake-like robots for search-and-rescue missions in cluttered environments, must rely on multiple degrees of freedom (DOFs). The inherent complexity associated with the coordinated control of a large number of actuators poses major difficulties and limitations in terms of structural intricacy and energy expenditure. This has become a crucial bottleneck in the development of modern robots, and therefore efforts are being invested to develop methods that enable controlling over several DOFs by means of a small number of inputs. The ideal visionary solution to this challenge would be devising a way to dictate any desired trajectory of a system comprised from several DOFs, with only one controlled input. We show that such fantastic ability, although seeming unrealistic at first glance, can be achieved using cleverly-designed multistable structures and exploiting their unique dynamic behavior. 

Currently, most dexterous robots considered in the literature, such as those mentioned above, are conventionally composed of rigid links connected by actuated joints \cite{lynch2017modern,niku2020introduction}. For interaction involving intermittent contact with objects and human-robot collaboration, the operation of such robots requires highly complicated control that involves force-sensing and coordination between multiple joints, as well as feedback stabilization. A promising remedy lies in the rapidly evolving field of \textit{soft robotics}, in which actuation is coupled with the elastic response of the robot’s structure to achieve complex continuous deformation patterns  \cite{shen2020stimuli,shepherd2011multigait,marchese2015recipe,tang2019programmable,lee2020hydrogel,cao2019arbitrarily,park2022hygroresponsive,palagi2016structured,rogoz2016light,da2020bioinspired,bar2002electroactive,wu2019insect,lum2016shape,wu2020multifunctional}. Soft actuators are utilizable for various robotic applications \cite{hines2017soft,el2020soft,coyle2018bio,pal2021exploiting,cao2021bistable}, such as bio-inspired legged locomotion \cite{shepherd2011multigait,venkiteswaran2019bio,goldfield2012bio}, aquatic locomotion \cite{marchese2014autonomous}, soft grippers \cite{https://doi.org/10.1002/adma.201707035,Wang}, deformable sheets \cite{Ren,petralia2010fabrication,wang2021integration}, and distributed sensing \cite{Zhang2018DistributedPO,REN2021103075,he2018multi}. Several actuation methods exist for soft actuators, such as control of flow rate or pressure in channel network or bladders embedded within an elastic structure \cite{morin2014using,morin2014elastomeric,bartlett20153d,mac2015poroelastic,de2005production}, electrostatic and piezo-electric activation of smart materials \cite{li2017fast,acome2018hydraulically,shintake2018soft,shahinpoor1998ionic,punning2011multilayer},  magnetic field \cite{mao2020soft,kim2018printing,hu2018small,fuhrer2009crosslinking,he2023magnetic}, thermal and thermo-chemical activation \cite{he2019electrically,zhao2022twisting,li2022three,stergiopulos2014soft,shepherd2013using,tolley2014untethered,wu2022fast,doi:10.1126/sciadv.adf8014}, and more. 

A crucial challenge in soft actuators is the typical need for multiple control inputs, which must be coordinated and involve closed-loop feedback and sensing. This may impose significant technical complications, limiting the realization of the control system. Moreover, accurate tracking of continuous time-varying deformation trajectories of an inertial-elastic structure using feedback control may also require high energy expenditure. 
A key concept which has been introduced for actuation of soft structures is the use of \textit{bi-stable elements}. A bi-stable element is characterized by two distinguished stable equilibrium configurations for the same prescribed load, separated by an unstable (spinodal) equilibrium state. Such behavior can be observed in curved beams \cite{vangbo1998analytical,zhao2008post,mises1923stabilitatsprobleme,arena2017adaptive,timoshenko2009theory}, thin-walled hyper-elastic balloons \cite{alexander1971tensile,muller2004rubber,overvelde2015amplifying,ben2020single}, and pre-stressed elastic sheets \cite{brodland1987deflection,holmes2007snapping,faber2020dome,taffetani2018static,guest2006analytical,armon2011geometry,gomez2017passive,liu2023snap,shui2022aligned,keplinger2012harnessing,dong2018near}.
An actuator made of multiple inter-connected bi-stable elements can transition between multiple stable states using minimal activation for snapping each element, while the rest of the motion is done passively \cite{kim2013soft,majidi2014soft,rus2015design,cianchetti2018biomedical,cianchetti2014soft,lee2017soft,chi2022bistable,katz2018solitary}. This enables achieving complex motions using minimal actuation energy by exploiting the multi-stable energy landscape of the system  \cite{tang2020leveraging,chen2018harnessing,baumgartner2020lesson,jiao2019advanced,zhakypov2019designing,treml2018origami,preston2019digital,glozman2010self}. For example, an actuator containing $N$ bi-stable elements, can provide a multitude of possible trajectories that involve transitions between $2^N$ stable states. This concept has been exploited in fluid-actuated soft robotic actuators \cite{overvelde2015amplifying,che2018viscoelastic,tang2020leveraging,milana2022morphological}, electrically and magnetically-actuated actuators \cite{gude2011piezoelectrically,medina2017latching,hou2018magneto}, for applications such as energy harvesting \cite{harne2013concise,betts2012optimal}, robotics \cite{gorissen2020inflatable,wang2023untethered,kaarthik2022motorized,zolfagharian2020closed}, MEMS \cite{wang2013constant}, energy absorption \cite{shan2015multistable}, metamaterials \cite{ma2019origami,wu2018dial}, 
folding of actuated origami sheets \cite{novelino2020untethered,kaufmann2022harnessing,son20224d,gillman2018design,yasuda2017origami}, 
and more. The main limitation of this concept is the need to actuate each of the bi-stable elements by a separate controlled input, which is still complicated.

In order to minimize the complexity of the control system for multi-stable actuators, the concept of \textit{single-input control} has been recently introduced. According to this concept, the bi-stable elements are interconnected and mechanically coupled. Moreover, the elements differ in their mechanical properties in a pre-designed way that enables dictating the order of transitions between multi-stable states of the actuator. This concept has been demonstrated in single-input control by \citet{gorissen2019hardware} and \citet{melancon2022inflatable}. In these works, the actuator was able to produce only specific sequences of state transitions. Also, \citet{che2018viscoelastic} proposed multi-material viscoelastic architected materials whose snapping sequence can be tuned using temperature as a control parameter. 
The work of \citet{ben2020single} presented single-input flow rate control of a chain of serially-connected hyper-elastic rubber balloons, where it was shown that it is possible, in principle, to reach all possible $2^N$ states. However, the \textit{sequences of transitions} from initial to target state were limited to a specific order, which may result in very long sequences. 
In a recent work, \citet{novelino2020untethered} presented an actuator consisting of serially-connected bi-stable cells with pre-tuned magnetization, which are actuated by an external magnetic field. While this actuator is able to transition directly between any desired stable states, it requires control of both the magnetic field's magnitude and direction, which are effectively two scalar inputs. 
Also, the use of high-energy magnetic fields is often not allowed in applications that involve sensitive electronics or optical components.

While all the actuators mentioned above rely on quasi-static changes of the states using slow input rate, we propose here a novel concept of \textit{exploiting the system's dynamics} in order to enable all possible transitions between neighboring multi-stable states using a single input. Moreover, careful design of the potential energy profiles of the mechanical bi-stable elements comprising the actuator, enables achieving added stable equilibrium states where one of the elements lies in its intermediate (spinodal) equilibrium. This enables increasing the number of multi-stable states for an $N$-element system up to $2^{N} (1+N/2)$, which gives a refined resolution of the actuator's states for the same number of elements.

\section{Results}
\begin{figure}[h!]
     \centering
     \includegraphics[width=0.95\textwidth]{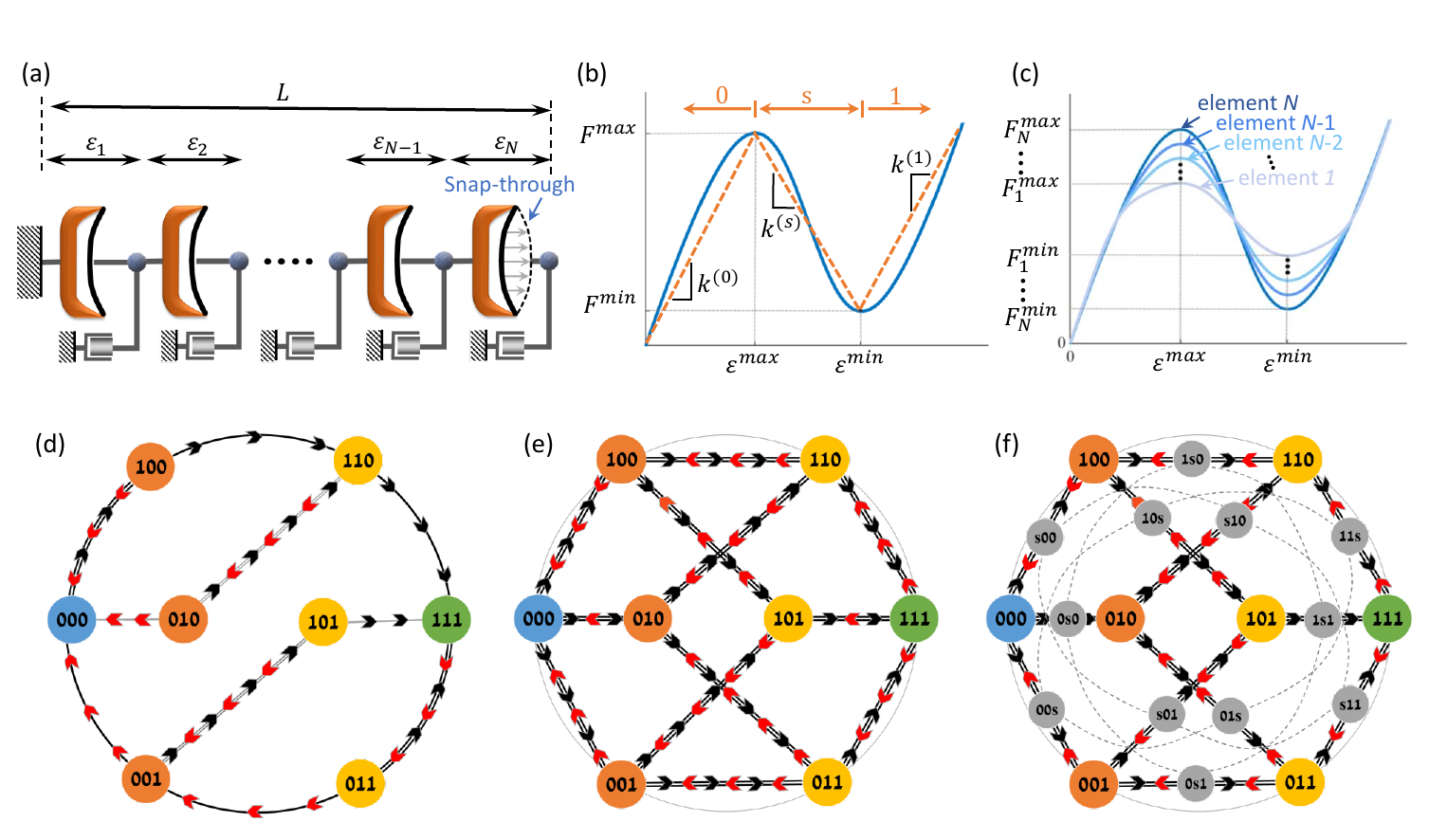}
    \caption{(a) Schematic illustration of a multi-stable chain, comprised from bi-stable elements that are connected in series. The endpoints of each element are also connected to dampers.  (b) The force-displacement relation $F(\varepsilon)$ of a bi-stable element is non-monotonic, with two states of positive stiffness, states `0' and `1', separated by a branch of negative stiffness termed the spinodal region or state `s'. The tri-linear approximation of this non-monotonic relation, denoted by the dashed straight-line segments, may offer useful analytical insights.
    (c) Pre-designed variability between the bi-stable elements which makes each element weaker or stronger compared to its neighbors, as in Eq.\,\eqref{eq:variability}. (d)-(f) Examples of several state transition diagrams for a multi-state actuator with $N=3$ bi-stable elements, which offers 8 (or $2^N$) possible binary states. Transitions between these states form a trajectory. (d) State-of-the-art single-input actuators with ordered variability or dissipative damping as in \citet{ben2020single} enable reaching any state, but the sequence of quasi-static transitions from initial to target state is limited to a specific order, which results in long, often impractical, sequences. Moreover, a complete cycle, from (000) to (111) and back to (000), is possible only through a single trajectory, thus termed single-input single-cycle actuators. (e) Single-input multiple-transition actuator. Our concept enables following any desired complete-cycle trajectory with a single input. The number of complete-cycle trajectories is 36 for $N=3$ (generally $ (N!)^2 $). Transitions occur by snap-through of an element from one stable phase to the other via an intermediate `s' phase of unstable equilibrium (spinodal), thus providing rapid transition. (f) High-resolution single-input multiple-transition actuator, where intermediate states having one element  at `s' phase are deliberately designed to be stable, enabling enhanced trajectory resolution with $20$ stable states for $N=3$ (generally $2^{N-1}(N+2)$).}
    \label{fig:Illustration}
\end{figure}

\subsection{Achieving single-input control of multi-state actuator}
Consider a chain of $N$ bi-stable elements connected in series, as illustrated in Fig.\,\ref{fig:Illustration}(a). Each bi-stable element is characterized by a non-monotonic force-displacement relation $F(\varepsilon)$ having two phases of positive stiffness separated by an intermediate branch of negative stiffness, see Fig.\,\ref{fig:Illustration}(b). In force-control conditions, the branch of negative stiffness, termed \textit{spinodal phase}, is unstable, while the two phases of positive stiffness are stable. Thus, practically, two distinctly separated equilibrium phases are possible, which we denote by `0' and `1', while the intermediate phase is denoted by `s'. Accordingly, the chain of bi-stable elements may be viewed as an actuator with $2^{N}$ stable equilibrium states, each corresponding to a different arrangement of zeros and ones, represented as a binary number. The sequence of transitions between these different binary states defines a trajectory. 
For example, with $N=3$, state $(111)$ may be reached through $N!=6$ different trajectories that begin at state $(000)$, and vice-versa, as illustrated in Fig.\,\ref{fig:Illustration}(e), giving rise to a total of $(N!)^2=36$ different complete-cycle trajectories. 
We propose to control such a multi-state actuator, i.e. to make it follow any desired trajectory, using a single input that controls the total length of the chain, $L(t)$, or particularly, the rate at which its length is changing, $\dot{L}=v(t)$. The concept is based on exploiting two competing symmetry-breaking mechanisms that, together, dictate the next transition event from a given state. This tug-of-war is decided by the rate $v(t)$, which is the only control input. Above, the notion of symmetry refers to the case where all bi-stable elements are identical, i.e., have the same force-displacement relation. Subjecting such a chain to quasi-static displacement-control conditions would lead to a random sequence of transition events since all elements reach the critical transition force together, that is $F^{max}$ during  extension or $F^{min}$ during  contraction. 

Breaking this symmetry is achieved by utilizing two different mechanisms. First, we introduce dissipative damping into the elements' motion, which adds rate-dependent forces. Second, we introduce pre-designed variability between the bi-stable elements in an ordered way, which is formulated as
\begin{equation} \label{eq:variability}
    F_i^{max}<F_{i+1}^{max} \; \mbox{ and } \; F_i^{min}> F_{i+1}^{min} \;\; \mbox{ for } i=1 \ldots N-1, 
\end{equation}
as illustrated schematically in Fig.\,\ref{fig:Illustration}(c). While the variability breaks the symmetry by making the elements "weaker" or "stronger", damping makes the actual force experienced by each of the bi-stable elements different from that of its neighbors. By clever design, we make these two mechanisms compete such that the input rate $v(t)$ dictates the level of dominance between them, and by that, enforce a desired order of transitions. 
Moreover, intermediate states, where one of the bi-stable elements is in the spinodal region (`s' phase), can be made stable or unstable by tuning the relations between elements' local stiffnesses. 
Creating stable intermediate states (for example (0\text{s}1)) enhance the resolution of the discretized trajectory for the same number of elements, $N$, as shown in Fig.\,\ref{fig:Illustration}(f). On the other hand, unstable intermediate states allow for rapid state transitions. Each of these two features may be advantageous, depending on the desired motion behavior for specific application, providing another level of tunability. 

\subsection{Experimental results}
\begin{figure}[h!]
     \centering
     \includegraphics[width=0.8\textwidth]{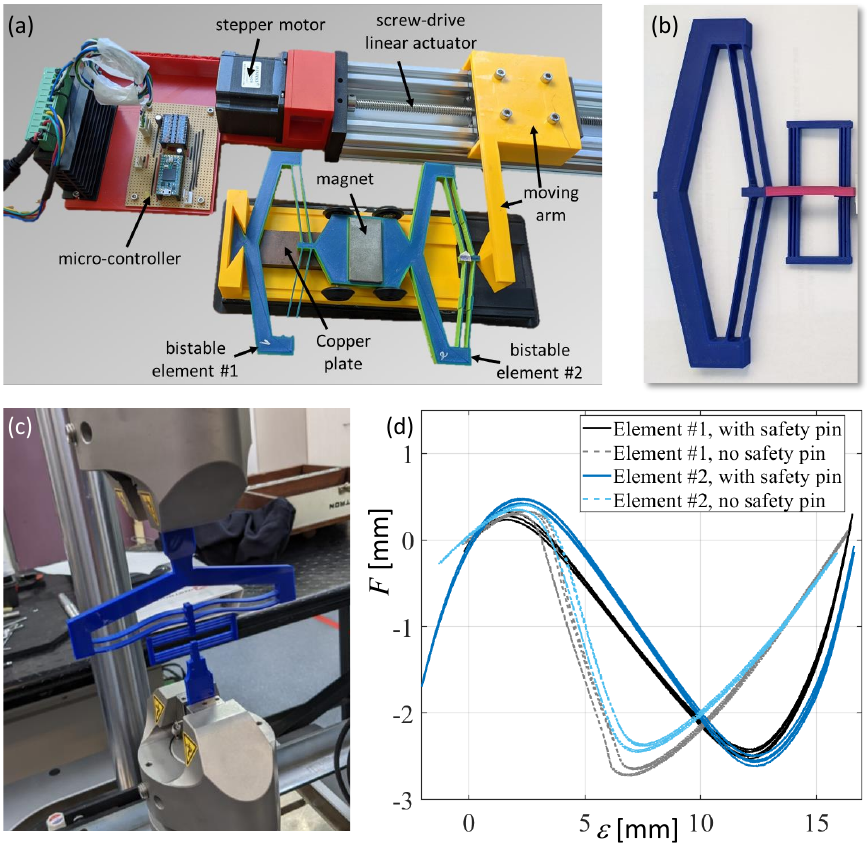}
      \caption{A multi-stable actuator with two bi-stable elements, $N=2$.  (a) Photo of our experimental setup. (b) Each bi-stable element is a 3-D printed double curved-beam structure. Variability between the force-displacement relations of the two bi-stable elements was introduced by means of differences in the geometry of the curved beams. Also, removal of the "safety pin" activates the serially-connected element acting as a linear spring, which changes the equivalent stiffness of phase `s' and therefore alters the stability of intermediate states. (c) The element with unlocked linear spring, clamped to Instron device for conducing force-displacement measurements. (d) Measurements of the bi-stable force-displacement profile for each of the two elements, with and without the safety pin, are shown for several  extension-contraction cycles, demonstrating high repeatability and small hysteresis.}
        \label{fig:twoElements}
\end{figure}

\begin{figure}[h!]
     \centering
     \includegraphics[width=1.0\textwidth]{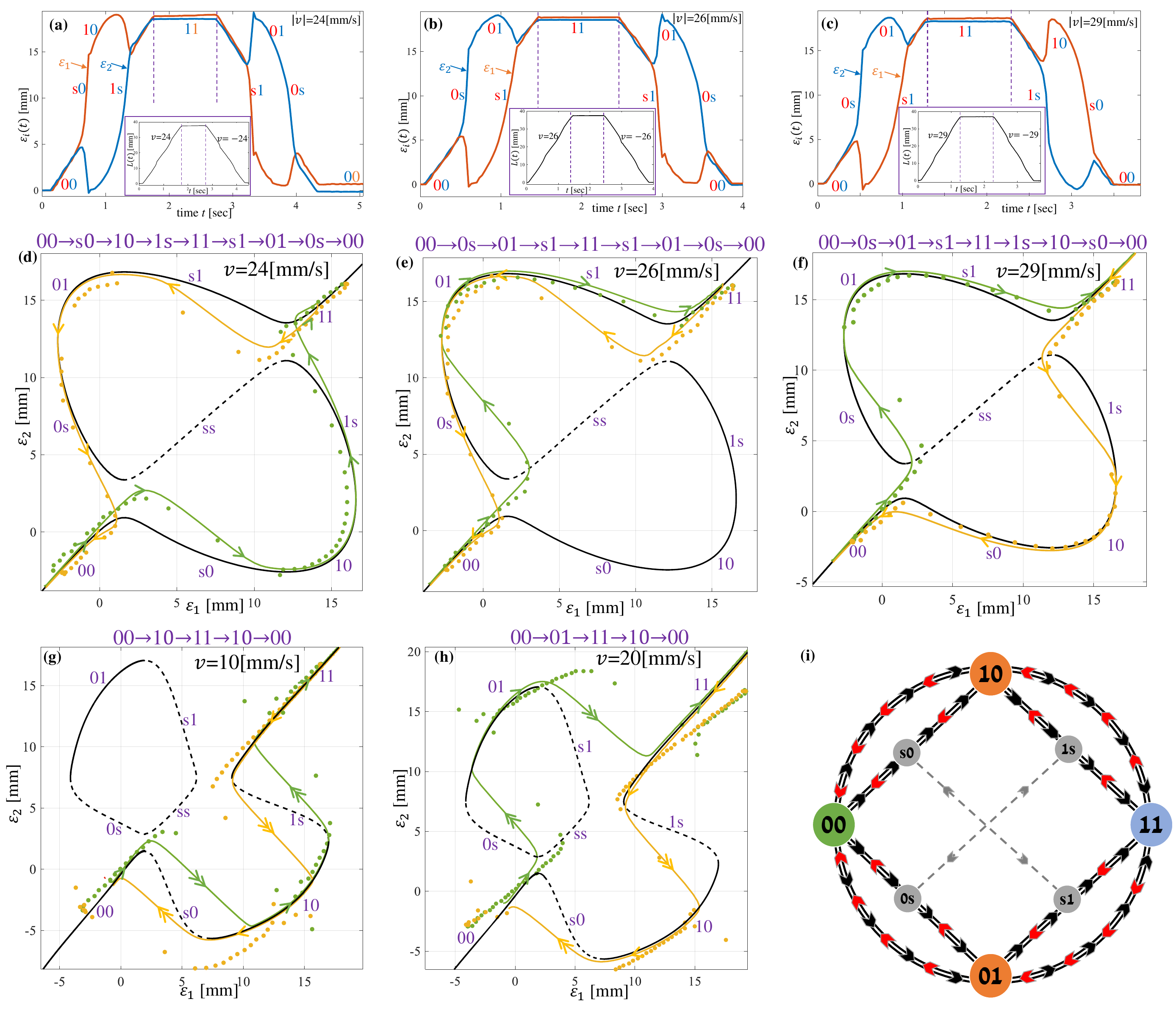}
        \caption{Experimental results of the two-element actuator. (a)-(c) Measured elongations of the elements $\varepsilon_i$ versus time $t$, for input rates $v$=\{24, 26, 29\}mm/s, respectively. (d)-(f) Measurements of  displacements plotted as dots in the plane of $(\varepsilon_1,\varepsilon_2)$, for input rates $v$=\{24, 26, 29\}mm/s, respectively. The three different trajectories demonstrate the ability to follow each of the sequences of state transitions by merely controlling the rate of the input (extension/contraction) rate $v$. For reference, the equilibrium curves (solid and dashed black curves for stable or unstable equilibrium, respectively) and the predictions of the numerical simulations of our theoretical dynamic model (orange and green solid curves with arrows) are illustrated as well. 
        (g)-(h) Experiment results with unstable intermediate states (without the safety pin). Note the abrupt transitions (snap-through) between binary states, denoted by double-arrows, in contrast to the smooth transition through intermediate states observed in (d)-(e)-(f). (i) State transition diagram of a two-element actuator with stable intermediate states.}      
        \label{fig:Exp2Elements}
\end{figure}

\begin{figure}[h!]
     \centering
     \includegraphics[width=0.85\textwidth]{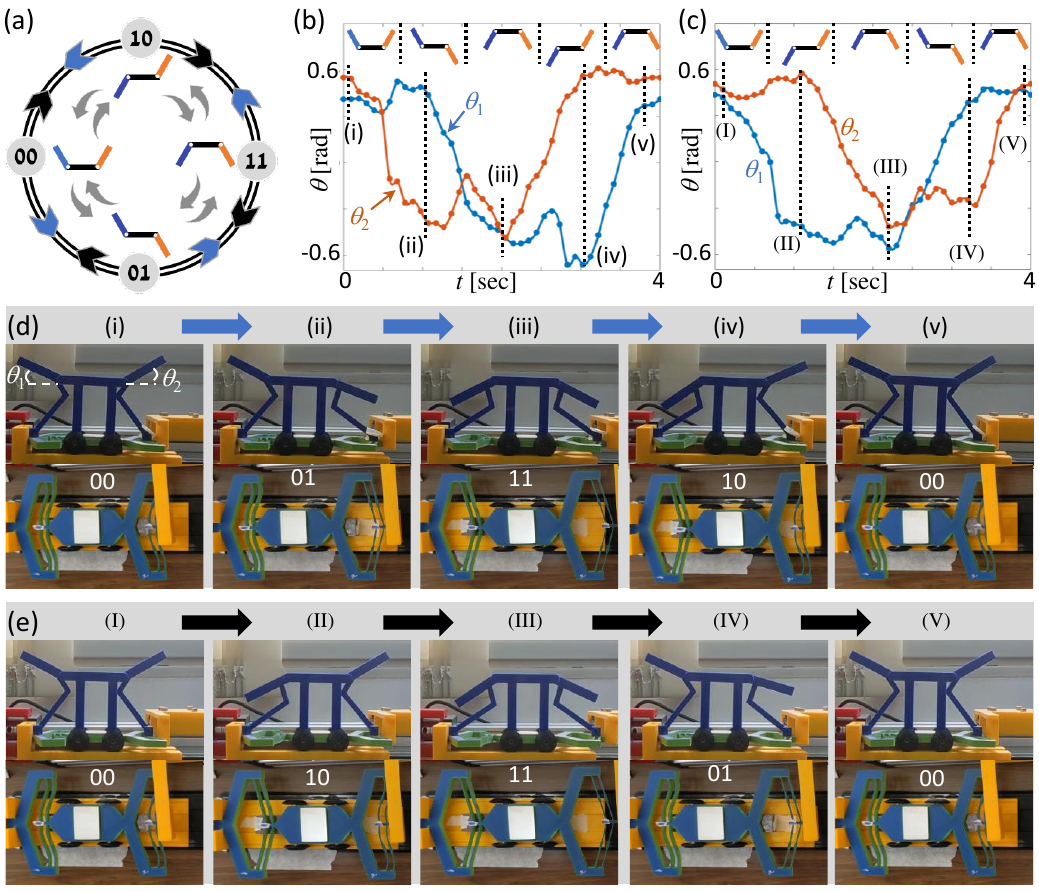}
      \caption{A 3-link mechanism controlled by a single input using the proposed actuator, allowing for any desired sequence of strokes. (a) A transitions diagram between binary states of the three-link mechanism. (b) Plots of measured the mechanism's two joint angles $\theta_i$ versus time $t$ for low input rate $|v|=20$mm/s, leading to state sequence $\{(00)\to (01)\to (11) \to (10) \to (00)\}$. (c) Plots of measured the mechanism's two joint angles $\theta_i$ versus time $t$ for low input rate $|v|=28$mm/s, leading to state sequence $\{(00)\to (10)\to (11) \to (01) \to (00)\}$;  Snapshot pictures from the movie of the mechanism's motion (d) for low input rate, (e) for high input rate. Below each snapshot sequence, we show illustrations of the bi-stable states of the actuator's two elements.}
      \label{fig:3link}
\end{figure}

In order to demonstrate the concept discussed above, we begin by describing the results of experiments involving a multi-stable actuator with two bi-stable elements, $N=2$. Here, the number of binary states is 4, $\{(00), (10), (01), (11)\}$, the number of intermediate states is 4, $\{(0\text{s}), (\text{s}0), (1\text{s}), (\text{s}1)\}$, and the number of possible trajectories in each direction (from state (00) to (11) or vice-versa) is two, giving rise to an overall of four different complete-cycle trajectories, see Fig.\,\ref{fig:Exp2Elements}(i). Each bi-stable element is made of a double curved-beam structure \cite{hussein2019design,salinas2015can}, see Fig.\,\ref{fig:twoElements}(b), which is manufactured by 3-D printing from PLA (see more details in section \cref{sec:MaM}). Importantly, the bi-stable behavior of the curved-beam element, e.g., its stiffness in different states or the level of forces $F^{max}$ and $F^{min}$, can be tuned by a careful design of the curved-beam geometry. Thus, variability between the force-displacement relations of the two bi-stable elements was introduced by means of differences in the geometry of the curved-beams (see section \cref{sec:MaM}), and the corresponding measured force-displacement relations are shown in Fig.\,\ref{fig:twoElements}(d).  
The ordered variability is dictated by Eq.\,\eqref{eq:variability}, as illustrated in Fig.\,\ref{fig:Illustration}(c). The two bi-stable elements are mounted on wheeled carts that move along a rail, see Fig.\,\ref{fig:twoElements}(a). Rate-dependent damping, taking the role of the dashpots in Fig.\,\ref{fig:Illustration}(a), is produced by attaching magnets to the bottom of the carts, which interact, without contact, with a copper plate that is fixed to the rail, see Fig.\,\ref{fig:twoElements}(a). The relative motion of the magnets with respect to the copper plate induces an electric field that creates eddy currents in the plate. In turn, a damping force that is proportional to the relative velocity is produced \cite{cadwell1996magnetic}. 

Figs.\,\ref{fig:3link}(a)-(f) 
show the results of  extension-contraction cycles performed at different input rates, demonstrating the ability to follow each of the four possible trajectories of state transitions by merely controlling the input rates (see also \href{https://technionmail-my.sharepoint.com/:v:/g/personal/givli_technion_ac_il/EakLq-ni-dBItjpPN3OYdT4BxnVbHNP0Hrb9WUjUxhnuwQ?e=9KMVSH}{the supplementary video}). As expected, at low rates, the weaker element changes its state before the stronger element. On the other hand, if the input rate is high enough, the damping forces become dominant, causing the stronger element to switch first by changing the distribution of forces along the chain. 
In these experiments, we designed the geometry of the bi-stable elements such that the intermediate states $\{(0\text{s}),(1\text{s}),(\text{s}0),(\text{s}1)\}$ are stable, which adds higher resolution of intermediate stable equilibrium states, as well as  possible transitions between them as illustrated in Fig.\,\ref{fig:Exp2Elements}(i). These intermediate states can be made unstable by increasing the relative stiffness of the spinodal region for each element. One way to achieve this is by connecting a linear spring in series to the bi-stable curved-beam element (see section \cref{sec:MaM}). Such a design is shown in Fig.\,\ref{fig:twoElements}(b)-(c), where the linear spring is realized by means of parallel bending beams, similar to a folded-beam structure often used in MEMS  \cite{shmulevich2014folded}.
When the "safety pin" is latched, as in the above-mentioned experiments, the linear-spring element is inactive, and the bi-stable behavior is dictated only by the curved-beam element. On the other hand, if the safety pin is removed, the linear spring is unlocked, which changes the overall stiffness of the bi-stable element such that the intermediate states become unstable. Such a simple locking mechanism, which determines whether the intermediate states are stable or unstable, may be useful for practical purposes. Moreover, since the serial connection of linear spring does not influence $F^{max}$ and $F^{min}$, the input rate required to follow each of the trajectories is identical to that of the previous case (with stable intermediate states).
The main difference between the two cases is the dynamic response in the intermediate states. If they are stable, the system moves quasi-statically, through a continuum of near-equilibrium states, as shown in Fig.\,\ref{fig:Exp2Elements}(a)-(f), and the response of the actuator is smooth and gradual even when a change of state occurs. On the other hand, if the intermediate states are unstable, any change of state results in an abrupt transition (snap through) to the next stable binary state, as illustrated in Fig.\,\ref{fig:Exp2Elements}(g)-(h). 

Since it is possible to follow any of the four complete-cycle trajectories by a single input, the multi-stable mechanism can be harnessed to function as an actuator that allows achieving four different cycles with a single-input control. This is demonstrated by connecting a three-link mechanism to the multi-stable actuator  with two elements, as shown in Fig.\,\ref{fig:3link}. The three-link mechanism is widely used as a basic model of planar locomotion of micro-swimmers \cite{becker2003self}, robotic snake swimmers in ideal fluid \cite{kanso2005locomotion}, and inchworm-like crawling \cite{gamus2020dynamic}. The important capability to generate locomotion relies on non-reversible undulatory motion of the two joint angles $\theta_1,\theta_2$ of the mechanism in phase-shifted periodic oscillations, which form a discretization of a traveling wave.
In contrast to the classical design which requires two separate inputs for controlling the joint angles, using our concept demonstrates that the choice of sequence of strokes of the three-link mechanism can be completely controlled by a single input (see also  \href{https://technionmail-my.sharepoint.com/:v:/g/personal/givli_technion_ac_il/EakLq-ni-dBItjpPN3OYdT4BxnVbHNP0Hrb9WUjUxhnuwQ?e=9KMVSH}{the supplementary video}). In particular, it is possible to generate a non-reciprocal "traveling wave" cycle in either direction, by appropriately choosing the rate of the single input, as shown in Fig.\,\ref{fig:3link}. This idea can be directly generalized to a larger number of degrees of freedom (DOFs), as discussed next.

\begin{figure}[t!]
     \centering
     \includegraphics[width=0.99\textwidth]{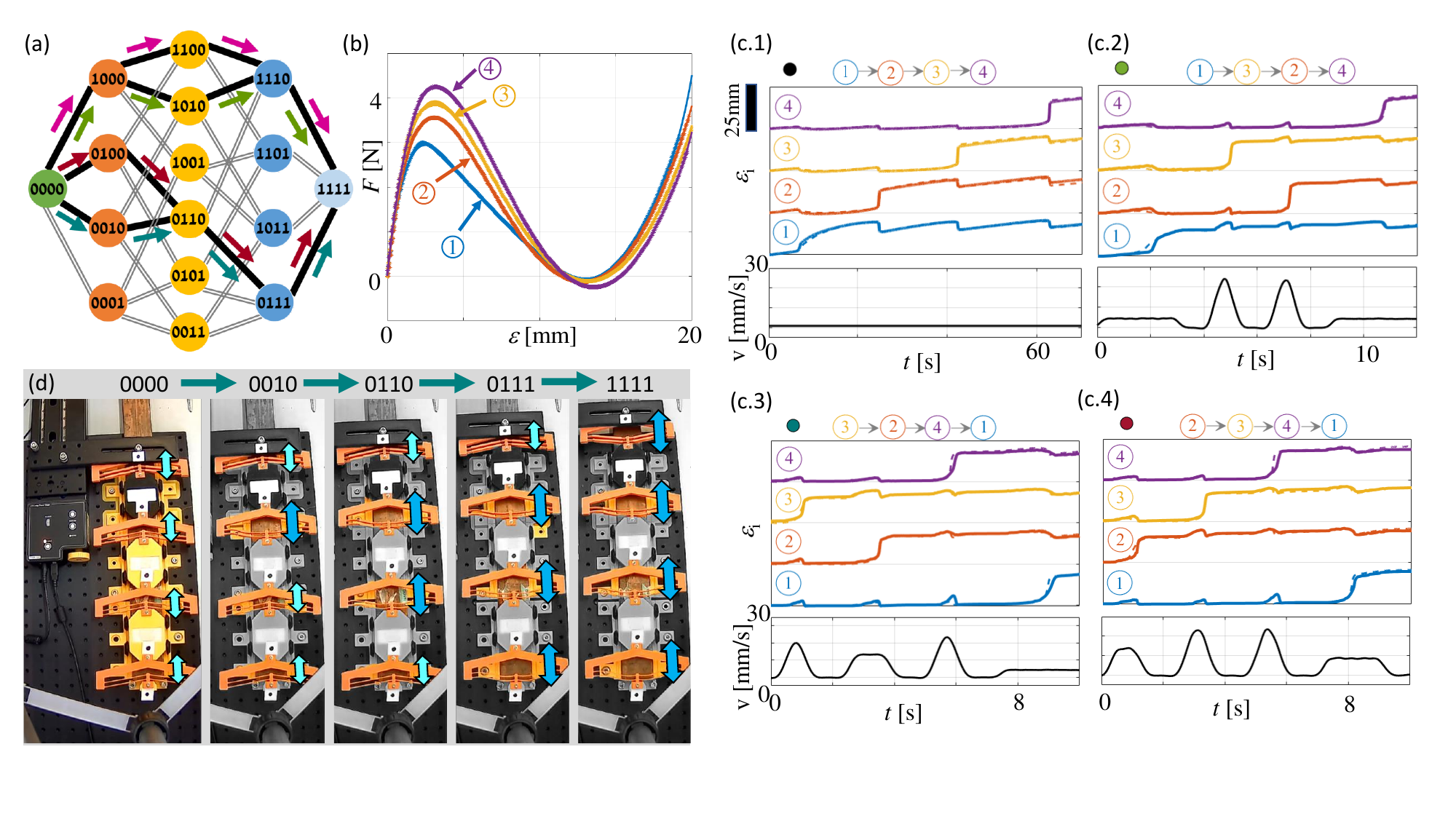}
        \caption{A multi-stable actuator with four bi-stable elements, $N=4$: 
        (a) Diagram with all possible binary states and transitions (not including intermediate `s' states). Each colored sequence of arrows indicates a different order of the snapping sequence, obtained experimentally. 
        (b) The measured force-displacement profiles of the four elements, which were designed with ordered variability. 
        (c) Plots of four time responses, each associated with a different experiment and a different snap-thorough sequence (see colored arrows in (a) corresponding to the colored circles in (c.1) through (c.4) ), showing the elongation,  $\varepsilon_{i}(t)$, of each of the bi-stable elements in a complete extension from state (0000) to state (1111). 
        different choice of time-varying input rate of extension $v(t)$, shown in the bottom graphs. Thus, each experiment results in a different trajectory with a chosen order of elements' snapping. (d) Snapshots from the experiment for which the order of snapping elements is $\{ 3\rightarrow 2 \rightarrow 4 \rightarrow 1\}$. That is, the sequence of binary state transitions is (0000)$\rightarrow$(0010)$\rightarrow$(0110)$\rightarrow$(0111)$\rightarrow$(1111). The small and large arrows in the pictures mark the closed and snapped-open elements, respectively. }
        \label{fig:4elements}
\end{figure}

\subsubsection{Experiments with four elements}
Fig.\,\ref{fig:4elements} shows the results of experiments for a multi-stable actuator comprising four bi-stable elements ($N=4$). In this case, the number of binary states is 16, the number of intermediate states is 32, and the number of complete-cycle trajectories is 576. Similar to the previous set of experiments, the bi-stable elements are designed and fabricated with variability in their critical forces $ F^{min} $ and $ F^{max} $, and  connected as a serial chain, ordered from weaker to stronger, see Fig.\,\ref{fig:4elements}(b). 
The elements are connected to each other by carts, to which strong magnets are attached in order to produce rate-dependent dampening through interaction with the copper plates that are fixed to the rail, similar to the setup in Fig.\,\ref{fig:twoElements}(a). Since the number of possible trajectories is extremely large (see Fig.\,\ref{fig:4elements}(a)), we present in Fig.\,\ref{fig:4elements}(d) the results for experimentally achieving four representative trajectories with a single input; these experiments are also shown in  \href{https://technionmail-my.sharepoint.com/:v:/g/personal/givli_technion_ac_il/EakLq-ni-dBItjpPN3OYdT4BxnVbHNP0Hrb9WUjUxhnuwQ?e=9KMVSH}{the supplementary video}. Nevertheless, the ability to follow any other trajectory has been demonstrated both experimentally and numerically (in simulations), by choosing the appropriate input rate at every state transition. Each plot in Fig.\,\ref{fig:4elements}(c) shows the time response (elongation versus time) of individual elements in a complete extension stage from state (0000) to state (1111). The only difference between these experiments is the time-varying input rate $v(t)$ at which the system is actuated.
This induces a different order in which the elements are "snapping" in 0 $\rightarrow$ 1 transitions. After each transition, the input rate is changed in order to select the next element to snap.

\subsection{Theoretical modeling and analysis}
We now present theoretical model and analysis of the multi-state actuator, which enables understanding the fundamental principles of our concept, as well as insightful guidelines for its design. 
The multi-state actuator is modeled as a set of $N$ bi-stable elements connected in series, whose endpoints are attached to $N$ linear rate-dependent dampers, as illustrated in Fig.\,\ref{fig:Illustration}(a). The elastic force acting on each bi-stable element is dictated by its force-displacement relationship, $F_i(\varepsilon_i)$, as illustrated in  Fig.\,\ref{fig:Illustration}(b). The dynamic equations of motion describing the response of the system are given in vector form as:
\begin{equation}
    \label{EOM}
    \cfrac{d\mathrm{\boldsymbol{\varepsilon}}}{dt}=\boldsymbol{\mathrm{WF}}(\boldsymbol{\varepsilon})+\boldsymbol{\mathrm{v}},
\end{equation}
where $\boldsymbol{\varepsilon}=[\varepsilon_{1},\ldots,\varepsilon_{N}]^{T}$, $\boldsymbol{\mathrm{F}}(\boldsymbol{\varepsilon})=[F_1(\varepsilon_{1}),F_2(\varepsilon_{2}),\ldots,F_N(\varepsilon_{N})]^{T}$, and  $\boldsymbol{\mathrm{W}}$ is a tri-diagonal matrix which takes the form:

\begin{equation}
\label{W}
    \boldsymbol{\mathrm{W}}=\cfrac{1}{c}
    \begin{bmatrix}
    -1 & 1  & 0  &\dots & 0 \\
     1 & -2 & 1  &\dots & 0 \\
    \vdots & \vdots & \vdots & \ddots & \vdots \\
    0 & 0  & \dots & 1 & -1 
\end{bmatrix},
\end{equation}
where $c$ is the damping coefficient, and $\boldsymbol{\mathrm{v}}=[0,0,\ldots,0,v(t)]^{T}$ with $v(t)=\dot L(t)$ being the extension rate of the chain, which is the controlled input. 

This set of ordinary differential equations (ODEs) is inherently nonlinear due to the non-monotonic force-displacement relation of the bi-stable elements $F_i(\varepsilon_{i})$, which are fitted from experiments as $5^{th}$-degree polynomials. Thus, Eq.\,\eqref{EOM} can be integrated numerically using a standard ODE solver in order to obtain the time response. 
Still, it is desired to gain some analytical insights and intuition. To this end, we replace the bi-stable force-displacement relation with a tri-linear profile, as illustrated in Fig.\,\ref{fig:Illustration}(b). This approximation preserves the main features of the bi-stable behavior, namely two phases of positive stiffness separated by a branch of negative stiffness, yet transforms the equations of motion into a set of piecewise-linear ODEs. More specifically, the governing equations are linear as long as the actuator's state has not changed, and nonlinearity is merely introduced by the change of states \cite{BENICHOU201394,NITECKI2021104634}. This important simplification enables analytical treatment, as discussed below. 

Before addressing the more general problem, it is instructive to consider some simple scenarios. First, it is easy to show that static equilibrium under $v=\dot L =0$ requires that the forces in all bi-stable elements are identical, subject to the constraint $\varepsilon_{1}+\cdot\cdot\cdot+\varepsilon_{N}=L$. Note that if all elements have the same force-displacement relation (no symmetry breaking), all elements that possess the same phase (0, 1 or s) must have the same elongation $\varepsilon_{i}$. Yet, due to the non-monotonic force-displacement relations of the bi-stable elements, the equilibrium solution is not unique. In other words, several equilibrium solutions, corresponding to different states, may be possible for the same prescribed extension $L$. The stability of these equilibrium solutions is dictated by the corresponding constrained Hessian matrix \cite{puglisi2000mechanics,benichou2013application}. 
According to the analysis in \cite{puglisi2000mechanics}, binary states (i.e. composed of only phases `0' and `1') are always stable, while states involving more than one element in spinodal `s' phase are unstable. Intermediate states, with exactly one bi-stable element in `s' phase are stable if and only if the actuator's equivalent stiffness $K_{eq}$ is negative, which is formulated as the condition \cite{puglisi2000mechanics}: 
\begin{equation}
    K_{eq}=
    \bigg(\cfrac{1}{k_1}+ \cfrac{1}{k_2} \cdots +  \cfrac{1}{k_N}\bigg)^{-1}<0
    \label{eq:intm_stability}
\end{equation}
where $k_i=dF_i / d \varepsilon_i$ is the local stiffness of the $i^{th}$ element.

\subsubsection{Analysis of actuator dynamics with two bi-stable elements}
Next, we examine the behavior of the simplest multi-state actuator, namely a system with two bi-stable elements, $N=2$.
In this case, equilibrium states and solutions of the actuator's dynamic response can be easily visualized as curves in the plane of relative displacements $(\varepsilon_1,\varepsilon_2)$. 

\begin{figure}[h!]
     \centering
     \includegraphics[width=0.99\textwidth]{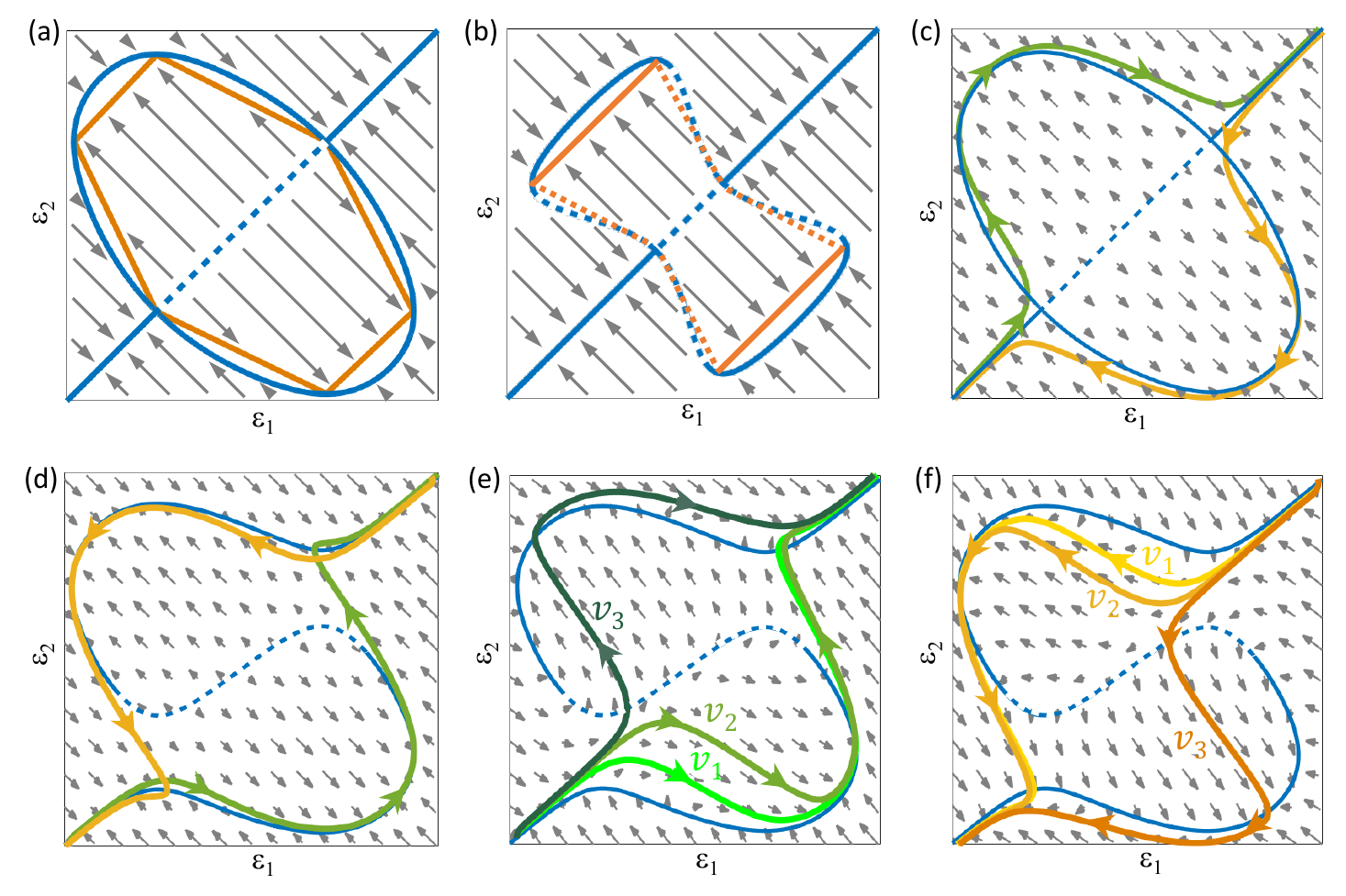}
        \caption{Illustrative maps of equilibrium curves in $(\varepsilon_1,\varepsilon_2)$ plane for multi-stable actuators with $N =2$ elements. The grid of arrows describes the vector field of solution trajectories, repelled from unstable equilibrium curves (dashed) towards stable equilibria (solid lines). (a) Two identical elements, with stable intermediate states $\{(0s),(s0),(1s),(s1)\}$. Blue curves correspond to polynomial profile $F(\varepsilon)$, whereas red line segments  correspond to its tri-linear approximation. (b) Case where the intermediate states are unstable equilibria. (c) The arrowed curves denote trajectories under small input rate, green color - extension $v>0$; orange color - contraction $v<0$. The damping forces break symmetry and enable a non-reversible cycle of state transitions sequence $(00)\to (01) \to (11) \to (00)$, as in \cite{ben2020single}. (d) With ordered variance between the elements as in \eqref{eq:variability}. The symmetry is broken and equilibrium curves change topology, such that element 1 snaps first in slow extension (green arrowed trajectory) and element 2 snaps first in slow contraction (orange arrowed trajectory). (e) Solution trajectories of extension $v>0$, starting from state $(00)$ with different input rates $v_1<v_2<v_c^+<v_3$. The high input rate $v_3$ which is beyond the critical rate $v_c^+$, changes the vector field $d\mathrm{\boldsymbol{{\varepsilon}}}/dt$ so that the trajectory can pass the "ridge line" of unstable equilibrium and fall into basin of attraction of state $(10)$, thus changing the order of snapping. (f) Solution trajectories of contraction $v<0$, starting from state $(11)$ with different input rates $|v_1|<|v_2|<|v_c^-|<|v_3|$. The highest contraction rate $v_3$ induces change in the order of snapping.   }
        \label{fig:model24elements}
\end{figure}

The equilibrium curves of a system with two identical bi-stable elements, $N=2$, having the same force-displacement relation $F(\varepsilon)$, are shown in Fig.\,\ref{fig:model24elements}(a). Unstable equilibrium points are denoted as dashed curves, whereas stable equilibria are shown as solid curves. The red straight-line segments represent equilibrium curves obtained while approximating $F(\varepsilon)$ by a tri-linear function, as illustrated in Fig.\,\ref{fig:Illustration}(b), where each straight segment represents a different state of the system (e.g.\,(01), (1s), etc.), being stable or unstable. The array of short arrows in Fig.\,\ref{fig:model24elements}(a) represent the vector field $d\mathrm{\boldsymbol{{\varepsilon}}}/dt=\boldsymbol{\mathrm{WF(\varepsilon)}}$ under zero input $v=0$ when starting from non-equilibrium initial values. 
Thus, solution trajectories move along lines of constant total length, $\varepsilon_1+\varepsilon_2=L_0$, i.e. in angle of $-45^\circ$, in a direction which is repelled from unstable equilibrium branches and attracted towards stable ones. The stability of equilibrium branches corresponding to intermediate states $\{(0s),(s0),(1s),(s1)\}$ is determined as follows. For tri-linear force-displacement relation, each phase has a constant stiffness $k^{(0)},k^{(1)}>0$ and $k^{(s)}<0$, and the stability condition in \eqref{eq:intm_stability} reduces to inequalities $k^{(0)},k^{(1)}>-k^{(s)}$. In the case of continuous (e.g. polynomial) force-displacement relation, the stiffnesses change continuously with $\varepsilon$, so that the branch associated with a given state can have stable and unstable parts. In the example shown in Fig.\,\ref{fig:model24elements}(b), the force-displacement relation $F(\varepsilon)$ is chosen such that condition \eqref{eq:intm_stability} is not satisfied, so that the branches of intermediate states are unstable (dashed lines/curves). If the systems in Fig.\,\ref{fig:model24elements}(a)-(b) move quasi-statically while slowly increasing the total length $L$  above $\varepsilon_{max}$ for leaving (00) state, the choice of transition to the next stable equilibrium state, i.e.,\,which elements is snapping first, is made at random, due to symmetry of the elements. However, for  small input velocity, the damping forces break this symmetry and enforces a choice of state transitions. This is illustrated in Fig.\,\ref{fig:model24elements}(c), which shows the equilibrium curves for the same identical elements as in Fig.\,\ref{fig:model24elements}(a), while the arrowed curves represent near-equilibrium solution trajectories of $\varepsilon_i(t)$ under small input velocity, $v>0$ for extension from (00) to (11) states, and $v<0$ for contraction back from (11) to (00), achieving a non-reversible cycle, as in \cite{ben2020single}. 

In order to enable a controlled choice between state transitions, we now introduce a second symmetry-breaking mechanism of \textit{designed variability} between the force-displacement relations of the elements, $F_i(\varepsilon_i)$. In particular, the variability is manifested by changing the critical forces $ F^{max} $ and $ F^{min} $ for each element, such that the second element is "weaker" than the first one both in  extension and  contraction, as given in Eq.\,\eqref{eq:variability}.  
The equilibrium curves for such case in $(\varepsilon_1,\varepsilon_2)$-plane appear in Fig.\,\ref{fig:model24elements}(d) (unstable branches in dashed curves), illustrating that the symmetry is broken.  
The colored arrowed curves represent solution trajectories under small input $v>0$ for extension from state (00), followed by $v<0$ for contraction back from (11). One can see that the trajectories are closely following the stable equilibrium curves, and then a rapid "snap" occurs when the stable equilibrium branch ends, where the solution is repelled and then attracted to a different stable branch. This illustrates a state transition sequence $(00)\to(10)\to(11)\to(01)\to(00)$.

Next, we consider the effect of increasing the input rate $v$. Assuming for simplicity that $v$ remains constant during extension, we show that there exists a critical rate $v_c^+$ such that for $v>v_c^+$ the state transition $(00)\to(10)$ changes to $(00) \to (01)$. This can be explained by visualizing the vector fields in $(\varepsilon_1,\varepsilon_2)$-plane. For very small input rate, the vector field is close to that in Fig.\,\ref{fig:model24elements}(d), and the solution trajectory follows closely the continuous stable equilibrium branch of $(00) \to (s0) \to(10)$. According to the system's dynamic equations \eqref{EOM}, increasing $v$ adds the constant upward vector $[0 , v]^T$ to the vector field at each point. Fig\,\ref{fig:model24elements}(e) shows solution trajectories $\varepsilon_i(t)$ under three different input rates $v_1<v_2<v_c^+<v_3$, where the vector field is shown for $v=v_3>v_c^+$. One can see how the vector field is rotated and drives the solution upwards, crossing the "gap" towards the stable branches of states $(0s)\to(01)$. A similar situation occurs in Fig.\,\ref{fig:model24elements}(f) for contraction from state (11) with three input rates $v=v_j<0$, such that $|v_1|<|v_2|<|v_c^-|<|v_3|$, with a different critical rate $v_c^-<0$. For $v=v_3$, the vector field has an added downward component which enables the trajectory to leave the (11) branch and cross the gap to the stable branches of states $(1s)\to(10)$. 

Next, we find explicit approximate expressions for the critical input rates $v_c^+,v_c^-$. This is done under some simplifying assumptions. First, we assume that the two elements have a tri-linear force-displacement relation, as in Fig.\,\ref{fig:Illustration}(b). That is, in state (00), the elastic force acting on each element is $F_i(\varepsilon_i)=k_i \varepsilon_i$. Second, we assume that the variability between elements changes their critical forces $F_i^{max}$, while the corresponding displacements $\varepsilon_i^{max}$ are kept equal, as illustrated in Fig.\,\ref{fig:Illustration}(c). This implies that $k_i=F_i^{max}/\varepsilon^{max}$. Using these assumptions, the solution of Eq.\,\eqref{EOM} for $\varepsilon_i(t)$ under zero initial conditions $\varepsilon_i=0$ is given by: 
\begin{equation} \label{eq:EOM2}
    \begin{array}{l}
    \varepsilon_1(t)=\left( \frac{k_2}{k_1+k_2}t - \frac{ck_2}{(k_1+k_2)^2} \left( 1-\mathrm{exp}(-\frac{k_1+k_2}{c}t) \right) \right)v, \\[15pt]
    \varepsilon_2(t)=\left( \frac{k_1}{k_1+k_2}t + \frac{ck_2}{(k_1+k_2)^2} \left( 1-\mathrm{exp}(-\frac{k_1+k_2}{c}t) \right) \right)v.
    \end{array}
\end{equation}
The element which reaches its critical force $F_i^{max}$ at displacement $\varepsilon_i=\varepsilon_{max}$ snaps first. For low input rate $v$, element 1 snaps first, and for larger rate $v>v_c^+$, element 2 snaps first. At the critical input rate $v=v_c^+$, both elements snap together at a time $t=t_c$, that is, $\varepsilon_1(t_c)=\varepsilon_2(t_c)=\varepsilon_{max}$. Substituting into \eqref{eq:EOM2} and eliminating $t_c$ gives an implicit transcendental equation for $v_c^+$, which is given in nondimensional form as:
\begin{equation} \label{eq:vc}
    \begin{array}{l}
        \tilde{v}_c^+ \left[1-\mathrm{exp} \left( -\dfrac{(1+\beta^{-1})(2-\alpha)+(1+\beta)(2+\alpha)}{2\tilde{v}_c^+} \right) \right]=\alpha(1+\beta);\qquad\quad\mbox{where } \\[15pt]
         \tilde{v}_c^+=\dfrac{c}{\bar F}v_c^+ ,\bar F = \dfrac{F_2^{max}+F_1^{max}}{2},\;\; \Delta F= F_2^{max}-F_1^{max} , \;\;  \beta=\dfrac{F_1^{max}}{F_2^{max}}<1 ,\;\; \alpha=\dfrac{\Delta F}{\bar F}=2 \dfrac{1+\beta}{1-\beta}.
    \end{array}
\end{equation}
Further assuming small variability between the elements, which implies $\alpha \ll 1$ and $\beta \to 1$, Eq.\,\eqref{eq:vc} can be asymptotically simplified to 
\begin{equation} \label{eq:asym_vc}
    \tilde{v}_c^+ = 2(1-\beta).
\end{equation}
A similar derivation can be made for the critical input rate in contraction for $v<0$, when moving from state (11), as follows. Assuming that the stiffnesses $k_i$ for each element in phase `1' are taken as equal to those of phase `0' under tri-linear profile (see Fig.\,\ref{fig:Illustration}(b)), the nondeimensional expression for critical input rate $\tilde{v}_c^-<0$ in contraction is similar to that in Eq.\,\eqref{eq:asym_vc}, with the modification $\beta=1-(F_1^{min}-F_2^{min})/{F_2^{max}}$. Table \ref{tab:vc} 
gives a comparison of the critical input rates $v_c$ for $v>0$ and $v<0$, calculated from \eqref{eq:asym_vc} and \eqref{eq:vc} under tri-linear approximation, as well as from numerical integration of \eqref{EOM} assuming polynomial profiles for $F_i(\varepsilon_i)$, as fitted from experiments. The last row in Table \ref{tab:vc} gives the actual critical input rates as obtained from experiments shown in Fig.\,\ref{fig:Exp2Elements}. 

\begin{table}[t!]
\centering
 \begin{tabular}{||c ||c || c | c||} 
 \hline
No. & critical input rate $|v_c|$   & Extension $v>0$ & Contraction $v<0$ \\ [0.5ex] 
 \hline\hline  
1. & Asymptotic approximation from \eqref{eq:asym_vc}          & 26.17 & 27.51 \\  [1ex] 
 2. & Analytical approximation from \eqref{eq:vc}          & 26.44 & 27.79 \\  [1ex] 
3. &  Integration of \eqref{EOM} for polynomial profile            & 25.59 & 26.30 \\ [1ex] 
4. &  Experimental results               & 25$\pm$1 & 27.5$\pm$1.5 \\  [1ex] 
 \hline
 \end{tabular}
 \caption{Magnitudes of the critical input rate $|v_c|$ in  $[mm / s]$ for the two-elements system considered in Fig.\,\ref{fig:twoElements}. 
 Row 1 - Using tri-linear functions for approximating measured profiles $F_i(\varepsilon_i)$ and the asymptotic approximation in Eq.\,\eqref{eq:asym_vc}. Row 2 - Using tri-linear functions and solving the transcendental equation \eqref{eq:vc}. Row 3 - Using fitted polynmial profiles $F_i(\varepsilon_i)$ and iterative numerical integration of the nonlinear dynamic equations \eqref{EOM}. Row 4 - actual values from experimental measurements for iteratively obtaining changes in state transition, with standard deviations indicated. The results demonstrate the predictive power of the theoretical model and its simplified approximations.   }
 \label{tab:vc}
\end{table}

\subsubsection{N elements}
By similar considerations as those applied in the analysis of a system with two bi-stable elements, one can directly generalize the concept to a multi-stable actuator with $ N $ elements, illustrated schematically in Fig.\,\ref{fig:Illustration}(a). To this end, pre-designed variability is introduced to the bi-stable behavior of the elements, and the elements are arranged in series from weaker to stronger, as formulated in Eq.\,\eqref{eq:variability} and illustrated in Fig.\,\ref{fig:Illustration}(c).
Consider a given stable equilibrium state (for example, binary state (0100) for $N=4$). Under very low input rate of extnesion $v>0$, all elements experience (almost) the same level of force; thus, the weakest element is the first to snap and change its phase from `0' to `1'. (in the example, it is element $i=1$). For higher input rate, the distribution of forces becomes less uniform, adding rate-dependent terms which become larger for elements located closer to the chain's actuated end. Thus, there is a critical rate above which the next element ($i=3$ in the example) snaps first to phase `1', and a higher rate that can make a farther element to snap first. The same principle applies also for contraction input $v<0$, and for stable intermediate states (e.g. (01s0)). That is, every transition between two stable states, as those shown in the state diagrams in Figs.\,\ref{fig:Illustration}(a),\ref{fig:Exp2Elements}(i),\ref{fig:4elements}(a) can be associated with an interval of input rates $v$ under which this transition is achieved. For $N=2$, the critical rates can be approximated using Eq.\,\eqref{eq:asym_vc}. For $N>2$, such analysis becomes more complicated, and the critical input rates which determine each state transition can be obtained using numerical intergation of the actuator's nonlinear dynamic equations \eqref{EOM} using fitted polynomial profiles for $F_i(\varepsilon_i)$, as done in our experimental demonstration with $N=4$ shown in Fig.\,\ref{fig:4elements}.    

\section{Materials and Methods}
\label{sec:MaM}
\subsection{Fabrication of bi-stable elements}
The bi-stable elements were made of PLA and fabricated using a 3D-printer Ultimaker 5s. Each element consists of two parallel pairs of clamped-clamped bending beams with locally varying thicknesses, as shown in Fig.\,\ref{fig:twoElements}(b). The geometry and thickness profiles of the beams were designed in order to create the desired bi-stable force-displacement profiles. 
In order to change the stiffness profile of the element and affect the stability of mixed multi-element states (such as `01s', etc.), each bi-stable element was connected in series to a secondary element comprised of six pairs of symmetry uniform beams bending about their symmetric configuration. This secondary element serves as a serially-connected linear spring having nearly-constant stiffness. Adding the secondary element affects the overall stiffness of the element without changing the critical forces for bi-stable state transitions. The secondary element has been equipped with a latch that can lock it rigidly in order to cancel its action and retain the element's original bi-stable stiffness profile, as shown in Fig.\,\ref{fig:twoElements}(b). When the latch is open, the secondary 'spring' is activated.

\subsection{Measurements of the bi-stable stiffness profile}

In order to characterize the force-displacement relation of each bi-stable element, we have applied standard elongation and contraction tests using Instron measurement system, as shown in Fig.\,\ref{fig:twoElements}(c). The output of the system gives data points of force measurements versus displacement. The discrete points obtained from measurements under four extension-contraction cycles, were then fitted by a 5$^{\text{th}}$-order polynomial, as shown in Fig.\,\ref{fig:twoElements}(d). The polynomial fit has also been used to extract the values at extremum points $\{\varepsilon_{min},\varepsilon_{max},F_{min},F_{max}\}$, which were then used in order to approximate the tri-linear profile, as shown in Fig.\,\ref{fig:Illustration}(b).

\subsection{Experimental setup}
The bi-stable elements have been serially connected to 3D-printed wheeled carts, moving on a supporting rail. In order to obtain large viscous damping, the cart has been equipped with a strong magnetic block, while an elongated copper plate (thickness 13mm) has been mounted to the rail, under the carts. The moving endpoint of the chain of elements has been connected to an arm that moves on a track, actuated by a lead screw connected to a stepper motor (HANPOSE 23H65628), which is commanded by a micro-controller, in order to obtain fixed-rate elongation or contraction of a chain of $N$ elements. The motion of the elements and the arm have been captured using a digital  camera mounted on top of the setup. Marker stickers on the elements were detected and tracked using Matlab image processing toolbox, in order to obtain time series of the elongation of each element. 

\subsection{Three-link mechanism}
The three-link mechanism has been mounted on the cart connecting two bi-stable elements, as shown in Fig.\,\ref{fig:3link}. Each side link of the mechanism has been connected to a bi-stable element by a ‘<’-shaped coupling link connected at its endpoints to passive joints. The dimensions of the coupling link and location of its connection points were designed such that the {open, closed} states of the bi-stable element result in elbow {up, down} angles of the mechanism’s side links, respectively. The mechanism’s links and their connecting joints were made of ABS, and 3D-printed using Ultimaker 5s printer. 

\subsection{Numerical analysis of the system's dynamics}
We have created a computational simulation code for the numerical integration of the N-element system’s coupled nonlinear dynamic equations \eqref{EOM}. The integration was performed in Matlab using the `ode45' adaptive-step integrator. The force-displacement relations of the simulated elements were based on the 5th-degree polynomials fitted from experimental measurements. The value of the damping coefficient in the experiments was extracted from preliminary calibration experiments of magnetic force measurements, and was estimated as $c=15$Ns/m. The numerical code enables analyzing equilibrium states of the N-element systems, as well as their stability. In addition, it enables numerically finding critical values of input rate of the total chain elongation/contraction that result in changing the order of state transitions, as well as their dependence on the bi-stability profiles of the elements. We have used the simulation code in order to design properties of our N-element systems for achieving desired sequences of state transitions.

\section{Discussion and conclusions}

We presented a novel concept that paves the way for realizing a new class of robotic actuation and sets the stage for the next-generation of soft robots with minimal control and maximal dexterity. This is accomplished by exploiting the unique dynamic behavior of multi-stable elastic structures that are comprised from an array of mechanically coupled bi-stable elements. By combining rate-dependent dissipation at the single-element level with carefully and cleverly pre-designed differences in the bi-stable behavior of the elements, we are able to facilitate any desired multi-state transition using a \textit{single} dynamic input. Our experiments showed that we can choose between any desired trajectory out of 576 possible ones with a 4-DOFs system, using a single control input. The actual use of such single-input control actuator was demonstrated by connecting it to a 3-link mechanism, representing a Purcell’s swimmer or a simple snake-like robot, where it was illustrated that it provides any desired functional deformation cycle with a single input. Importantly, the proposed concept can be realized at any practical scale and it does not require activation by external (e.g. magnetic or electric) fields. The latter is particularly important since the employment of such fields may restrict the use of the actuator due to potential interference with sensitive electronics or optical components. Our extensive analysis, based on theoretical, numerical and experimental study, has validated the feasibility and applicability of this concept, and provided important insights and guidelines regarding the design of such multi-state actuators. It was also shown that the extraordinary ability to choose any desired sequence of state-transitions can be directly extended to any number of DOFs, enabling the control over a very large number of possible trajectories, which goes as $(N!)^2$ where $N$ is the number of DOFs, with a \textit{single} controlled input.

\section*{Acknowledgements}
S.G. acknowledges support from the Israel Science Foundation grant 1598/21. The work of Y.O. and A.G. was supported by the PMRI – Peter Munk Research Institute – Technion, under grant no. 2031886, and the Technion Autonomous Systems Program under grant 2031975.
\newpage
\bibliographystyle{unsrtnat}
\bibliography{references}  

\begin{thebibliography}{124}
\providecommand{\natexlab}[1]{#1}
\providecommand{\url}[1]{\texttt{#1}}
\expandafter\ifx\csname urlstyle\endcsname\relax
  \providecommand{\doi}[1]{doi: #1}\else
  \providecommand{\doi}{doi: \begingroup \urlstyle{rm}\Url}\fi

\bibitem[Lynch and Park(2017)]{lynch2017modern}
Kevin~M Lynch and Frank~C Park.
\newblock \emph{Modern robotics}.
\newblock Cambridge University Press, 2017.

\bibitem[Niku(2020)]{niku2020introduction}
Saeed~B Niku.
\newblock \emph{Introduction to robotics: analysis, control, applications}.
\newblock John Wiley \& Sons, 2020.

\bibitem[Shen et~al.(2020)Shen, Chen, Zhu, Yong, and Gu]{shen2020stimuli}
Zequn Shen, Feifei Chen, Xiangyang Zhu, Ken-Tye Yong, and Guoying Gu.
\newblock Stimuli-responsive functional materials for soft robotics.
\newblock \emph{Journal of Materials Chemistry B}, 8\penalty0 (39):\penalty0 8972--8991, 2020.

\bibitem[Shepherd et~al.(2011)Shepherd, Ilievski, Choi, Morin, Stokes, Mazzeo, Chen, Wang, and Whitesides]{shepherd2011multigait}
Robert~F Shepherd, Filip Ilievski, Wonjae Choi, Stephen~A Morin, Adam~A Stokes, Aaron~D Mazzeo, Xin Chen, Michael Wang, and George~M Whitesides.
\newblock Multigait soft robot.
\newblock \emph{Proceedings of the national academy of sciences}, 108\penalty0 (51):\penalty0 20400--20403, 2011.

\bibitem[Marchese et~al.(2015)Marchese, Katzschmann, and Rus]{marchese2015recipe}
Andrew~D Marchese, Robert~K Katzschmann, and Daniela Rus.
\newblock A recipe for soft fluidic elastomer robots.
\newblock \emph{Soft robotics}, 2\penalty0 (1):\penalty0 7--25, 2015.

\bibitem[Tang et~al.(2019)Tang, Li, Hong, Yang, and Yin]{tang2019programmable}
Yichao Tang, Yanbin Li, Yaoye Hong, Shu Yang, and Jie Yin.
\newblock Programmable active kirigami metasheets with more freedom of actuation.
\newblock \emph{Proceedings of the National Academy of Sciences}, 116\penalty0 (52):\penalty0 26407--26413, 2019.

\bibitem[Lee et~al.(2020)Lee, Song, and Sun]{lee2020hydrogel}
Y~Lee, WJ~Song, and J-Y Sun.
\newblock Hydrogel soft robotics.
\newblock \emph{Materials Today Physics}, 15:\penalty0 100258, 2020.

\bibitem[Cao et~al.(2019)Cao, Zhou, Su, Zhang, Zhou, Zhou, and Yang]{cao2019arbitrarily}
Jie Cao, Changlin Zhou, Gehong Su, Xinxing Zhang, Tao Zhou, Zehang Zhou, and Yibo Yang.
\newblock Arbitrarily 3d configurable hygroscopic robots with a covalent--noncovalent interpenetrating network and self-healing ability.
\newblock \emph{Advanced Materials}, 31\penalty0 (18):\penalty0 1900042, 2019.

\bibitem[Park et~al.(2022)Park, Ha, Shin, and Kim]{park2022hygroresponsive}
Keunhwan Park, Jonghyun Ha, Beomjune Shin, and Ho-Young Kim.
\newblock Hygroresponsive movements of plants and soft actuators.
\newblock \emph{Soft Matter in Plants: From Biophysics to Biomimetics}, 2022.

\bibitem[Palagi et~al.(2016)Palagi, Mark, Reigh, Melde, Qiu, Zeng, Parmeggiani, Martella, Sanchez-Castillo, Kapernaum, et~al.]{palagi2016structured}
Stefano Palagi, Andrew~G Mark, Shang~Yik Reigh, Kai Melde, Tian Qiu, Hao Zeng, Camilla Parmeggiani, Daniele Martella, Alberto Sanchez-Castillo, Nadia Kapernaum, et~al.
\newblock Structured light enables biomimetic swimming and versatile locomotion of photoresponsive soft microrobots.
\newblock \emph{Nature materials}, 15\penalty0 (6):\penalty0 647--653, 2016.

\bibitem[Rog{\'o}{\.z} et~al.(2016)Rog{\'o}{\.z}, Zeng, Xuan, Wiersma, and Wasylczyk]{rogoz2016light}
Miko{\l}aj Rog{\'o}{\.z}, Hao Zeng, Chen Xuan, Diederik~Sybolt Wiersma, and Piotr Wasylczyk.
\newblock Light-driven soft robot mimics caterpillar locomotion in natural scale.
\newblock \emph{Advanced Optical Materials}, 4\penalty0 (11):\penalty0 1689--1694, 2016.

\bibitem[Da~Cunha et~al.(2020)Da~Cunha, Debije, and Schenning]{da2020bioinspired}
M~Pilz Da~Cunha, Michael~G Debije, and Albert~PHJ Schenning.
\newblock Bioinspired light-driven soft robots based on liquid crystal polymers.
\newblock \emph{Chemical Society Reviews}, 49\penalty0 (18):\penalty0 6568--6578, 2020.

\bibitem[Bar-Cohen(2002)]{bar2002electroactive}
Yoseph Bar-Cohen.
\newblock Electroactive polymers as artificial muscles: a review.
\newblock \emph{Journal of Spacecraft and Rockets}, 39\penalty0 (6):\penalty0 822--827, 2002.

\bibitem[Wu et~al.(2019)Wu, Yim, Liang, Shao, Qi, Zhong, Luo, Yan, Zhang, Wang, et~al.]{wu2019insect}
Yichuan Wu, Justin~K Yim, Jiaming Liang, Zhichun Shao, Mingjing Qi, Junwen Zhong, Zihao Luo, Xiaojun Yan, Min Zhang, Xiaohao Wang, et~al.
\newblock Insect-scale fast moving and ultrarobust soft robot.
\newblock \emph{Science Robotics}, 4\penalty0 (32):\penalty0 eaax1594, 2019.

\bibitem[Lum et~al.(2016)Lum, Ye, Dong, Marvi, Erin, Hu, and Sitti]{lum2016shape}
Guo~Zhan Lum, Zhou Ye, Xiaoguang Dong, Hamid Marvi, Onder Erin, Wenqi Hu, and Metin Sitti.
\newblock Shape-programmable magnetic soft matter.
\newblock \emph{Proceedings of the National Academy of Sciences}, 113\penalty0 (41):\penalty0 E6007--E6015, 2016.

\bibitem[Wu et~al.(2020)Wu, Hu, Ze, Sitti, and Zhao]{wu2020multifunctional}
Shuai Wu, Wenqi Hu, Qiji Ze, Metin Sitti, and Ruike Zhao.
\newblock Multifunctional magnetic soft composites: A review.
\newblock \emph{Multifunctional materials}, 3\penalty0 (4):\penalty0 042003, 2020.

\bibitem[Hines et~al.(2017)Hines, Petersen, Lum, and Sitti]{hines2017soft}
Lindsey Hines, Kirstin Petersen, Guo~Zhan Lum, and Metin Sitti.
\newblock Soft actuators for small-scale robotics.
\newblock \emph{Advanced materials}, 29\penalty0 (13):\penalty0 1603483, 2017.

\bibitem[El-Atab et~al.(2020)El-Atab, Mishra, Al-Modaf, Joharji, Alsharif, Alamoudi, Diaz, Qaiser, and Hussain]{el2020soft}
Nazek El-Atab, Rishabh~B Mishra, Fhad Al-Modaf, Lana Joharji, Aljohara~A Alsharif, Haneen Alamoudi, Marlon Diaz, Nadeem Qaiser, and Muhammad~Mustafa Hussain.
\newblock Soft actuators for soft robotic applications: a review.
\newblock \emph{Advanced Intelligent Systems}, 2\penalty0 (10):\penalty0 2000128, 2020.

\bibitem[Coyle et~al.(2018)Coyle, Majidi, LeDuc, and Hsia]{coyle2018bio}
Stephen Coyle, Carmel Majidi, Philip LeDuc, and K~Jimmy Hsia.
\newblock Bio-inspired soft robotics: Material selection, actuation, and design.
\newblock \emph{Extreme Mechanics Letters}, 22:\penalty0 51--59, 2018.

\bibitem[Pal et~al.(2021)Pal, Restrepo, Goswami, and Martinez]{pal2021exploiting}
Aniket Pal, Vanessa Restrepo, Debkalpa Goswami, and Ramses~V Martinez.
\newblock Exploiting mechanical instabilities in soft robotics: control, sensing, and actuation.
\newblock \emph{Advanced Materials}, 33\penalty0 (19):\penalty0 2006939, 2021.

\bibitem[Cao et~al.(2021)Cao, Derakhshani, Fang, Huang, and Cao]{cao2021bistable}
Yunteng Cao, Masoud Derakhshani, Yuhui Fang, Guoliang Huang, and Changyong Cao.
\newblock Bistable structures for advanced functional systems.
\newblock \emph{Advanced Functional Materials}, 31\penalty0 (45):\penalty0 2106231, 2021.

\bibitem[Venkiteswaran et~al.(2019)Venkiteswaran, Samaniego, Sikorski, and Misra]{venkiteswaran2019bio}
Venkatasubramanian~Kalpathy Venkiteswaran, Luis Fernando~Pena Samaniego, Jakub Sikorski, and Sarthak Misra.
\newblock Bio-inspired terrestrial motion of magnetic soft millirobots.
\newblock \emph{IEEE Robotics and automation letters}, 4\penalty0 (2):\penalty0 1753--1759, 2019.

\bibitem[Goldfield et~al.(2012)Goldfield, Park, Chen, Hsu, Young, Wehner, Kelty-Stephen, Stirling, Weinberg, Newman, et~al.]{goldfield2012bio}
Eugene~C Goldfield, Yong-Lae Park, Bor-Rong Chen, Wen-Hao Hsu, Diana Young, Michael Wehner, Damian~G Kelty-Stephen, Leia Stirling, Marc Weinberg, Dava Newman, et~al.
\newblock Bio-inspired design of soft robotic assistive devices: The interface of physics, biology, and behavior.
\newblock \emph{Ecological Psychology}, 24\penalty0 (4):\penalty0 300--327, 2012.

\bibitem[Marchese et~al.(2014)Marchese, Onal, and Rus]{marchese2014autonomous}
Andrew~D Marchese, Cagdas~D Onal, and Daniela Rus.
\newblock Autonomous soft robotic fish capable of escape maneuvers using fluidic elastomer actuators.
\newblock \emph{Soft robotics}, 1\penalty0 (1):\penalty0 75--87, 2014.

\bibitem[Shintake et~al.(2018{\natexlab{a}})Shintake, Cacucciolo, Floreano, and Shea]{https://doi.org/10.1002/adma.201707035}
Jun Shintake, Vito Cacucciolo, Dario Floreano, and Herbert Shea.
\newblock Soft robotic grippers.
\newblock \emph{Advanced Materials}, 30\penalty0 (29):\penalty0 1707035, 2018{\natexlab{a}}.
\newblock \doi{https://doi.org/10.1002/adma.201707035}.

\bibitem[Wang and Ahn(2017)]{Wang}
Wei Wang and Sung-Hoon Ahn.
\newblock Shape memory alloy-based soft gripper with variable stiffness for compliant and effective grasping.
\newblock \emph{Soft Robotics}, 4, 10 2017.
\newblock \doi{10.1089/soro.2016.0081}.

\bibitem[Ren et~al.(2021{\natexlab{a}})Ren, Zhang, Soon, Liu, Hu, Onck, and Sitti]{Ren}
Ziyu Ren, Rongjing Zhang, Ren Soon, Zemin Liu, Wenqi Hu, Patrick Onck, and Metin Sitti.
\newblock Soft-bodied adaptive multimodal locomotion strategies in fluid-filled confined spaces.
\newblock \emph{Science Advances}, 7:\penalty0 eabh2022, 06 2021{\natexlab{a}}.
\newblock \doi{10.1126/sciadv.abh2022}.

\bibitem[Petralia and Wood(2010)]{petralia2010fabrication}
Michael~T Petralia and Robert~J Wood.
\newblock Fabrication and analysis of dielectric-elastomer minimum-energy structures for highly-deformable soft robotic systems.
\newblock In \emph{2010 IEEE/RSJ International Conference on Intelligent Robots and Systems}, pages 2357--2363. IEEE, 2010.

\bibitem[Wang et~al.(2021)Wang, Wu, Huang, Du, Yue, Chen, Li, and Su]{wang2021integration}
Qi~Wang, Zhenhua Wu, Jianyu Huang, Zhuolin Du, Yamei Yue, Dezhi Chen, Dong Li, and Bin Su.
\newblock Integration of sensing and shape-deforming capabilities for a bioinspired soft robot.
\newblock \emph{Composites Part B: Engineering}, 223:\penalty0 109116, 2021.

\bibitem[Zhang and Polygerinos(2018)]{Zhang2018DistributedPO}
Wenlong Zhang and Panagiotis Polygerinos.
\newblock Distributed planning of multi-segment soft robotic arms.
\newblock \emph{2018 Annual American Control Conference (ACC)}, pages 2096--2101, 2018.

\bibitem[Ren et~al.(2021{\natexlab{b}})Ren, Li, Wei, Wang, Song, Wei, Ren, and {Qingping Liu}]{REN2021103075}
Luquan Ren, Bingqian Li, Guowu Wei, Kunyang Wang, Zhengyi Song, Yuyang Wei, Lei Ren, and {Qingping Liu}.
\newblock Biology and bioinspiration of soft robotics: Actuation, sensing, and system integration.
\newblock \emph{iScience}, 24\penalty0 (9):\penalty0 103075, 2021{\natexlab{b}}.
\newblock ISSN 2589-0042.
\newblock \doi{https://doi.org/10.1016/j.isci.2021.103075}.

\bibitem[He et~al.(2018)He, Zhou, Wang, Wang, Shen, and Wu]{he2018multi}
Bin He, Yanmin Zhou, Zhipeng Wang, Qigang Wang, Runjie Shen, and Shangqing Wu.
\newblock A multi-layered touch-pressure sensing ionogel material suitable for sensing integrated actuations of soft robots.
\newblock \emph{Sensors and Actuators A: Physical}, 272:\penalty0 341--348, 2018.

\bibitem[Morin et~al.(2014{\natexlab{a}})Morin, Shevchenko, Lessing, Kwok, Shepherd, Stokes, and Whitesides]{morin2014using}
Stephen~A Morin, Yanina Shevchenko, Joshua Lessing, Sen~Wai Kwok, Robert~F Shepherd, Adam~A Stokes, and George~M Whitesides.
\newblock Using “click-e-bricks” to make 3d elastomeric structures.
\newblock \emph{Advanced Materials}, 26\penalty0 (34):\penalty0 5991--5999, 2014{\natexlab{a}}.

\bibitem[Morin et~al.(2014{\natexlab{b}})Morin, Kwok, Lessing, Ting, Shepherd, Stokes, and Whitesides]{morin2014elastomeric}
Stephen~A Morin, Sen~Wai Kwok, Joshua Lessing, Jason Ting, Robert~F Shepherd, Adam~A Stokes, and George~M Whitesides.
\newblock Elastomeric tiles for the fabrication of inflatable structures.
\newblock \emph{Advanced Functional Materials}, 24\penalty0 (35):\penalty0 5541--5549, 2014{\natexlab{b}}.

\bibitem[Bartlett et~al.(2015)Bartlett, Tolley, Overvelde, Weaver, Mosadegh, Bertoldi, Whitesides, and Wood]{bartlett20153d}
Nicholas~W Bartlett, Michael~T Tolley, Johannes~TB Overvelde, James~C Weaver, Bobak Mosadegh, Katia Bertoldi, George~M Whitesides, and Robert~J Wood.
\newblock A 3d-printed, functionally graded soft robot powered by combustion.
\newblock \emph{Science}, 349\penalty0 (6244):\penalty0 161--165, 2015.

\bibitem[Mac~Murray et~al.(2015)Mac~Murray, An, Robinson, van Meerbeek, O'Brien, Zhao, and Shepherd]{mac2015poroelastic}
Benjamin~C Mac~Murray, Xintong An, Sanlin~S Robinson, Ilse~M van Meerbeek, Kevin~W O'Brien, Huichan Zhao, and Robert~F Shepherd.
\newblock Poroelastic foams for simple fabrication of complex soft robots.
\newblock \emph{Advanced Materials}, 27\penalty0 (41):\penalty0 6334--6340, 2015.

\bibitem[De~Volder et~al.(2005)De~Volder, Peirs, Reynaerts, Coosemans, Puers, Smal, and Raucent]{de2005production}
Micha{\"e}l De~Volder, Jan Peirs, Dominiek Reynaerts, Johan Coosemans, Robert Puers, Olivier Smal, and Benoit Raucent.
\newblock Production and characterization of a hydraulic microactuator.
\newblock \emph{Journal of Micromechanics and microengineering}, 15\penalty0 (7):\penalty0 S15, 2005.

\bibitem[Li et~al.(2017)Li, Li, Liang, Cheng, Dai, Yang, Liu, Zeng, Huang, Luo, et~al.]{li2017fast}
Tiefeng Li, Guorui Li, Yiming Liang, Tingyu Cheng, Jing Dai, Xuxu Yang, Bangyuan Liu, Zedong Zeng, Zhilong Huang, Yingwu Luo, et~al.
\newblock Fast-moving soft electronic fish.
\newblock \emph{Science advances}, 3\penalty0 (4):\penalty0 e1602045, 2017.

\bibitem[Acome et~al.(2018)Acome, Mitchell, Morrissey, Emmett, Benjamin, King, Radakovitz, and Keplinger]{acome2018hydraulically}
Eric Acome, Shane~K Mitchell, TG~Morrissey, MB~Emmett, Claire Benjamin, Madeline King, Miles Radakovitz, and Christoph Keplinger.
\newblock Hydraulically amplified self-healing electrostatic actuators with muscle-like performance.
\newblock \emph{Science}, 359\penalty0 (6371):\penalty0 61--65, 2018.

\bibitem[Shintake et~al.(2018{\natexlab{b}})Shintake, Cacucciolo, Floreano, and Shea]{shintake2018soft}
Jun Shintake, Vito Cacucciolo, Dario Floreano, and Herbert Shea.
\newblock Soft robotic grippers.
\newblock \emph{Advanced materials}, 30\penalty0 (29):\penalty0 1707035, 2018{\natexlab{b}}.

\bibitem[Shahinpoor et~al.(1998)Shahinpoor, Bar-Cohen, Simpson, and Smith]{shahinpoor1998ionic}
Mohsen Shahinpoor, Yoseph Bar-Cohen, JO~Simpson, and J~Smith.
\newblock Ionic polymer-metal composites (ipmcs) as biomimetic sensors, actuators and artificial muscles-a review.
\newblock \emph{Smart materials and structures}, 7\penalty0 (6):\penalty0 R15, 1998.

\bibitem[Punning et~al.(2011)Punning, Akbari, Niklaus, and Shea]{punning2011multilayer}
Andres Punning, Samin Akbari, Muhamed Niklaus, and Herbert Shea.
\newblock Multilayer dielectric elastomer actuators with ion implanted electrodes.
\newblock In \emph{Electroactive Polymer Actuators and Devices (EAPAD) 2011}, volume 7976, pages 259--266. SPIE, 2011.

\bibitem[Mao et~al.(2020)Mao, Drack, Karami-Mosammam, Wirthl, Stockinger, Schw{\"o}diauer, and Kaltenbrunner]{mao2020soft}
Guoyong Mao, Michael Drack, Mahya Karami-Mosammam, Daniela Wirthl, Thomas Stockinger, Reinhard Schw{\"o}diauer, and Martin Kaltenbrunner.
\newblock Soft electromagnetic actuators.
\newblock \emph{Science advances}, 6\penalty0 (26):\penalty0 eabc0251, 2020.

\bibitem[Kim et~al.(2018)Kim, Yuk, Zhao, Chester, and Zhao]{kim2018printing}
Yoonho Kim, Hyunwoo Yuk, Ruike Zhao, Shawn~A Chester, and Xuanhe Zhao.
\newblock Printing ferromagnetic domains for untethered fast-transforming soft materials.
\newblock \emph{Nature}, 558\penalty0 (7709):\penalty0 274--279, 2018.

\bibitem[Hu et~al.(2018)Hu, Lum, Mastrangeli, and Sitti]{hu2018small}
Wenqi Hu, Guo~Zhan Lum, Massimo Mastrangeli, and Metin Sitti.
\newblock Small-scale soft-bodied robot with multimodal locomotion.
\newblock \emph{Nature}, 554\penalty0 (7690):\penalty0 81--85, 2018.

\bibitem[Fuhrer et~al.(2009)Fuhrer, Athanassiou, Luechinger, and Stark]{fuhrer2009crosslinking}
Roland Fuhrer, Evagelos~Kimon Athanassiou, Norman~Albert Luechinger, and Wendelin~Jan Stark.
\newblock Crosslinking metal nanoparticles into the polymer backbone of hydrogels enables preparation of soft, magnetic field-driven actuators with muscle-like flexibility.
\newblock \emph{Small}, 5\penalty0 (3):\penalty0 383--388, 2009.

\bibitem[He et~al.(2023)He, Tang, Hu, Yang, Xu, Zr{\'\i}nyi, and Chen]{he2023magnetic}
Yuan He, Jie Tang, Yang Hu, Sen Yang, Feng Xu, Miklos Zr{\'\i}nyi, and Yong~Mei Chen.
\newblock Magnetic hydrogel-based flexible actuators: A comprehensive review on design, properties, and applications.
\newblock \emph{Chemical Engineering Journal}, 462:\penalty0 142193, 2023.

\bibitem[He et~al.(2019)He, Wang, Wang, Minori, Tolley, and Cai]{he2019electrically}
Qiguang He, Zhijian Wang, Yang Wang, Adriane Minori, Michael~T Tolley, and Shengqiang Cai.
\newblock Electrically controlled liquid crystal elastomer--based soft tubular actuator with multimodal actuation.
\newblock \emph{Science advances}, 5\penalty0 (10):\penalty0 eaax5746, 2019.

\bibitem[Zhao et~al.(2022)Zhao, Chi, Hong, Li, Yang, and Yin]{zhao2022twisting}
Yao Zhao, Yinding Chi, Yaoye Hong, Yanbin Li, Shu Yang, and Jie Yin.
\newblock Twisting for soft intelligent autonomous robot in unstructured environments.
\newblock \emph{Proceedings of the National Academy of Sciences}, 119\penalty0 (22):\penalty0 e2200265119, 2022.

\bibitem[Li et~al.(2022)Li, Teixeira, Parlato, Grace, Wang, Huey, and Wang]{li2022three}
Yi~Li, Yasmin Teixeira, Gina Parlato, Jaclyn Grace, Fei Wang, Bryan~D Huey, and Xueju Wang.
\newblock Three-dimensional thermochromic liquid crystal elastomer structures with reversible shape-morphing and color-changing capabilities for soft robotics.
\newblock \emph{Soft Matter}, 18\penalty0 (36):\penalty0 6857--6867, 2022.

\bibitem[Stergiopulos et~al.(2014)Stergiopulos, Vogt, Tolley, Wehner, Barber, Whitesides, and Wood]{stergiopulos2014soft}
Constantinos Stergiopulos, Daniel Vogt, Michael~T Tolley, Michael Wehner, Jabulani Barber, George~M Whitesides, and Robert~J Wood.
\newblock A soft combustion-driven pump for soft robots.
\newblock In \emph{Smart Materials, Adaptive Structures and Intelligent Systems}, volume 46155, page V002T04A011. American Society of Mechanical Engineers, 2014.

\bibitem[Shepherd et~al.(2013)Shepherd, Stokes, Freake, Barber, Snyder, Mazzeo, Cademartiri, Morin, and Whitesides]{shepherd2013using}
Robert~F Shepherd, Adam~A Stokes, Jacob Freake, Jabulani Barber, Phillip~W Snyder, Aaron~D Mazzeo, Ludovico Cademartiri, Stephen~A Morin, and George~M Whitesides.
\newblock Using explosions to power a soft robot.
\newblock \emph{Angewandte Chemie International Edition}, 52\penalty0 (10):\penalty0 2892--2896, 2013.

\bibitem[Tolley et~al.(2014)Tolley, Shepherd, Karpelson, Bartlett, Galloway, Wehner, Nunes, Whitesides, and Wood]{tolley2014untethered}
Michael~T Tolley, Robert~F Shepherd, Michael Karpelson, Nicholas~W Bartlett, Kevin~C Galloway, Michael Wehner, Rui Nunes, George~M Whitesides, and Robert~J Wood.
\newblock An untethered jumping soft robot.
\newblock In \emph{2014 IEEE/RSJ International Conference on Intelligent Robots and Systems}, pages 561--566. IEEE, 2014.

\bibitem[Wu et~al.(2022)Wu, Baker, Yin, and Zhu]{wu2022fast}
Shuang Wu, Gregory~Langston Baker, Jie Yin, and Yong Zhu.
\newblock Fast thermal actuators for soft robotics.
\newblock \emph{Soft Robotics}, 9\penalty0 (6):\penalty0 1031--1039, 2022.

\bibitem[Wu et~al.(2023)Wu, Hong, Zhao, Yin, and Zhu]{doi:10.1126/sciadv.adf8014}
Shuang Wu, Yaoye Hong, Yao Zhao, Jie Yin, and Yong Zhu.
\newblock Caterpillar-inspired soft crawling robot with distributed programmable thermal actuation.
\newblock \emph{Science Advances}, 9\penalty0 (12):\penalty0 eadf8014, 2023.

\bibitem[Vangbo(1998)]{vangbo1998analytical}
Mattias Vangbo.
\newblock An analytical analysis of a compressed bistable buckled beam.
\newblock \emph{Sensors and Actuators A: Physical}, 69\penalty0 (3):\penalty0 212--216, 1998.

\bibitem[Zhao et~al.(2008)Zhao, Jia, He, and Wang]{zhao2008post}
Jian Zhao, Jianyuan Jia, Xiaoping He, and Hongxi Wang.
\newblock Post-buckling and snap-through behavior of inclined slender beams.
\newblock \emph{Journal of Applied Mechanics}, 75\penalty0 (4), 2008.

\bibitem[Mises(1923)]{mises1923stabilitatsprobleme}
RV~Mises.
\newblock {\"U}ber die stabilit{\"a}tsprobleme der elastizit{\"a}tstheorie.
\newblock \emph{ZAMM-Journal of Applied Mathematics and Mechanics/Zeitschrift f{\"u}r Angewandte Mathematik und Mechanik}, 3\penalty0 (6):\penalty0 406--422, 1923.

\bibitem[Arena et~al.(2017)Arena, MJ~Groh, Brinkmeyer, Theunissen, M.~Weaver, and Pirrera]{arena2017adaptive}
Gaetano Arena, Rainer MJ~Groh, Alex Brinkmeyer, Raf Theunissen, Paul M.~Weaver, and Alberto Pirrera.
\newblock Adaptive compliant structures for flow regulation.
\newblock \emph{Proceedings of the Royal Society A: Mathematical, Physical and Engineering Sciences}, 473\penalty0 (2204):\penalty0 20170334, 2017.

\bibitem[Timoshenko and Gere(2009)]{timoshenko2009theory}
Stephen~P Timoshenko and James~M Gere.
\newblock \emph{Theory of elastic stability}.
\newblock Courier Corporation, 2009.

\bibitem[Alexander(1971)]{alexander1971tensile}
Harold Alexander.
\newblock Tensile instability of initially spherical balloons.
\newblock \emph{International Journal of Engineering Science}, 9\penalty0 (1):\penalty0 151--160, 1971.

\bibitem[M{\"u}ller and Strehlow(2004)]{muller2004rubber}
Ingo M{\"u}ller and Peter Strehlow.
\newblock \emph{Rubber and rubber balloons: paradigms of thermodynamics}, volume 637.
\newblock Springer Science \& Business Media, 2004.

\bibitem[Overvelde et~al.(2015)Overvelde, Kloek, D’haen, and Bertoldi]{overvelde2015amplifying}
Johannes~TB Overvelde, Tamara Kloek, Jonas~JA D’haen, and Katia Bertoldi.
\newblock Amplifying the response of soft actuators by harnessing snap-through instabilities.
\newblock \emph{Proceedings of the National Academy of Sciences}, 112\penalty0 (35):\penalty0 10863--10868, 2015.

\bibitem[Ben-Haim et~al.(2020)Ben-Haim, Salem, Or, and Gat]{ben2020single}
Eran Ben-Haim, Lior Salem, Yizhar Or, and Amir~D Gat.
\newblock Single-input control of multiple fluid-driven elastic actuators via interaction between bistability and viscosity.
\newblock \emph{Soft Robotics}, 7\penalty0 (2):\penalty0 259--265, 2020.

\bibitem[Brodland and Cohen(1987)]{brodland1987deflection}
GW~Brodland and H~Cohen.
\newblock Deflection and snapping of spherical caps.
\newblock \emph{International journal of solids and structures}, 23\penalty0 (10):\penalty0 1341--1356, 1987.

\bibitem[Holmes and Crosby(2007)]{holmes2007snapping}
Douglas~P Holmes and Alfred~J Crosby.
\newblock Snapping surfaces.
\newblock \emph{Advanced Materials}, 19\penalty0 (21):\penalty0 3589--3593, 2007.

\bibitem[Faber et~al.(2020)Faber, Udani, Riley, Studart, and Arrieta]{faber2020dome}
Jakob~A Faber, Janav~P Udani, Katherine~S Riley, Andr{\'e}~R Studart, and Andres~F Arrieta.
\newblock Dome-patterned metamaterial sheets.
\newblock \emph{Advanced Science}, 7\penalty0 (22):\penalty0 2001955, 2020.

\bibitem[Taffetani et~al.(2018)Taffetani, Jiang, Holmes, and Vella]{taffetani2018static}
Matteo Taffetani, Xin Jiang, Douglas~P Holmes, and Dominic Vella.
\newblock Static bistability of spherical caps.
\newblock \emph{Proceedings of the Royal Society A: Mathematical, Physical and Engineering Sciences}, 474\penalty0 (2213):\penalty0 20170910, 2018.

\bibitem[Guest and Pellegrino(2006)]{guest2006analytical}
SD~Guest and S~Pellegrino.
\newblock Analytical models for bistable cylindrical shells.
\newblock \emph{Proceedings of the Royal Society A: Mathematical, Physical and Engineering Sciences}, 462\penalty0 (2067):\penalty0 839--854, 2006.

\bibitem[Armon et~al.(2011)Armon, Efrati, Kupferman, and Sharon]{armon2011geometry}
Shahaf Armon, Efi Efrati, Raz Kupferman, and Eran Sharon.
\newblock Geometry and mechanics in the opening of chiral seed pods.
\newblock \emph{Science}, 333\penalty0 (6050):\penalty0 1726--1730, 2011.

\bibitem[Gomez et~al.(2017)Gomez, Moulton, and Vella]{gomez2017passive}
Michael Gomez, Derek~E Moulton, and Dominic Vella.
\newblock Passive control of viscous flow via elastic snap-through.
\newblock \emph{Physical review letters}, 119\penalty0 (14):\penalty0 144502, 2017.

\bibitem[Liu et~al.(2023)Liu, Domino, de~Dinechin, Taffetani, and Vella]{liu2023snap}
Mingchao Liu, Lucie Domino, Iris~Dupont de~Dinechin, Matteo Taffetani, and Dominic Vella.
\newblock Snap-induced morphing: From a single bistable shell to the origin of shape bifurcation in interacting shells.
\newblock \emph{Journal of the Mechanics and Physics of Solids}, 170:\penalty0 105116, 2023.

\bibitem[Shui et~al.(2022)Shui, Ni, and Wang]{shui2022aligned}
Langquan Shui, Ke~Ni, and Zhengzhi Wang.
\newblock Aligned magnetic nanocomposites for modularized and recyclable soft microrobots.
\newblock \emph{ACS Applied Materials \& Interfaces}, 14\penalty0 (38):\penalty0 43802--43814, 2022.

\bibitem[Keplinger et~al.(2012)Keplinger, Li, Baumgartner, Suo, and Bauer]{keplinger2012harnessing}
Christoph Keplinger, Tiefeng Li, Richard Baumgartner, Zhigang Suo, and Siegfried Bauer.
\newblock Harnessing snap-through instability in soft dielectrics to achieve giant voltage-triggered deformation.
\newblock \emph{Soft Matter}, 8\penalty0 (2):\penalty0 285--288, 2012.

\bibitem[Dong et~al.(2018)Dong, Tong, Zhang, Chen, and Zhao]{dong2018near}
Liangliang Dong, Xia Tong, Hongji Zhang, Mingqing Chen, and Yue Zhao.
\newblock Near-infrared light-driven locomotion of a liquid crystal polymer trilayer actuator.
\newblock \emph{Materials Chemistry Frontiers}, 2\penalty0 (7):\penalty0 1383--1388, 2018.

\bibitem[Kim et~al.(2013)Kim, Laschi, and Trimmer]{kim2013soft}
S~Kim, C~Laschi, and B~Trimmer.
\newblock Soft robotics: a new perspective in robot evolution.
\newblock \emph{Trends Biotechnol}, 31:\penalty0 287--294, 2013.

\bibitem[Majidi(2014)]{majidi2014soft}
Carmel Majidi.
\newblock Soft robotics: a perspective—current trends and prospects for the future.
\newblock \emph{Soft robotics}, 1\penalty0 (1):\penalty0 5--11, 2014.

\bibitem[Rus and Tolley(2015)]{rus2015design}
Daniela Rus and Michael~T Tolley.
\newblock Design, fabrication and control of soft robots.
\newblock \emph{Nature}, 521\penalty0 (7553):\penalty0 467--475, 2015.

\bibitem[Cianchetti et~al.(2018)Cianchetti, Laschi, Menciassi, and Dario]{cianchetti2018biomedical}
Matteo Cianchetti, Cecilia Laschi, Arianna Menciassi, and Paolo Dario.
\newblock Biomedical applications of soft robotics.
\newblock \emph{Nature Reviews Materials}, 3\penalty0 (6):\penalty0 143--153, 2018.

\bibitem[Cianchetti et~al.(2014)Cianchetti, Ranzani, Gerboni, Nanayakkara, Althoefer, Dasgupta, and Menciassi]{cianchetti2014soft}
Matteo Cianchetti, Tommaso Ranzani, Giada Gerboni, Thrishantha Nanayakkara, Kaspar Althoefer, Prokar Dasgupta, and Arianna Menciassi.
\newblock Soft robotics technologies to address shortcomings in today's minimally invasive surgery: the stiff-flop approach.
\newblock \emph{Soft robotics}, 1\penalty0 (2):\penalty0 122--131, 2014.

\bibitem[Lee et~al.(2017)Lee, Kim, Kim, Hong, Ryu, Kim, and Kim]{lee2017soft}
Chiwon Lee, Myungjoon Kim, Yoon~Jae Kim, Nhayoung Hong, Seungwan Ryu, H~Jin Kim, and Sungwan Kim.
\newblock Soft robot review.
\newblock \emph{International Journal of Control, Automation and Systems}, 15\penalty0 (1):\penalty0 3--15, 2017.

\bibitem[Chi et~al.(2022)Chi, Li, Zhao, Hong, Tang, and Yin]{chi2022bistable}
Yinding Chi, Yanbin Li, Yao Zhao, Yaoye Hong, Yichao Tang, and Jie Yin.
\newblock Bistable and multistable actuators for soft robots: Structures, materials, and functionalities.
\newblock \emph{Advanced Materials}, 34\penalty0 (19):\penalty0 2110384, 2022.

\bibitem[Katz and Givli(2018)]{katz2018solitary}
Shmuel Katz and Sefi Givli.
\newblock Solitary waves in a bistable lattice.
\newblock \emph{Extreme Mechanics Letters}, 22:\penalty0 106--111, 2018.

\bibitem[Tang et~al.(2020)Tang, Chi, Sun, Huang, Maghsoudi, Spence, Zhao, Su, and Yin]{tang2020leveraging}
Yichao Tang, Yinding Chi, Jiefeng Sun, Tzu-Hao Huang, Omid~H Maghsoudi, Andrew Spence, Jianguo Zhao, Hao Su, and Jie Yin.
\newblock Leveraging elastic instabilities for amplified performance: Spine-inspired high-speed and high-force soft robots.
\newblock \emph{Science advances}, 6\penalty0 (19):\penalty0 eaaz6912, 2020.

\bibitem[Chen et~al.(2018)Chen, Bilal, Shea, and Daraio]{chen2018harnessing}
Tian Chen, Osama~R Bilal, Kristina Shea, and Chiara Daraio.
\newblock Harnessing bistability for directional propulsion of soft, untethered robots.
\newblock \emph{Proceedings of the National Academy of Sciences}, 115\penalty0 (22):\penalty0 5698--5702, 2018.

\bibitem[Baumgartner et~al.(2020)Baumgartner, Kogler, Stadlbauer, Foo, Kaltseis, Baumgartner, Mao, Keplinger, Koh, Arnold, et~al.]{baumgartner2020lesson}
Richard Baumgartner, Alexander Kogler, Josef~M Stadlbauer, Choon~Chiang Foo, Rainer Kaltseis, Melanie Baumgartner, Guoyong Mao, Christoph Keplinger, Soo Jin~Adrian Koh, Nikita Arnold, et~al.
\newblock A lesson from plants: high-speed soft robotic actuators.
\newblock \emph{Advanced Science}, 7\penalty0 (5):\penalty0 1903391, 2020.

\bibitem[Jiao et~al.(2019)Jiao, Zhang, Wang, Pan, Yang, and Zou]{jiao2019advanced}
Zhongdong Jiao, Chao Zhang, Wei Wang, Min Pan, Huayong Yang, and Jun Zou.
\newblock Advanced artificial muscle for flexible material-based reconfigurable soft robots.
\newblock \emph{Advanced Science}, 6\penalty0 (21):\penalty0 1901371, 2019.

\bibitem[Zhakypov et~al.(2019)Zhakypov, Mori, Hosoda, and Paik]{zhakypov2019designing}
Zhenishbek Zhakypov, Kazuaki Mori, Koh Hosoda, and Jamie Paik.
\newblock Designing minimal and scalable insect-inspired multi-locomotion millirobots.
\newblock \emph{Nature}, 571\penalty0 (7765):\penalty0 381--386, 2019.

\bibitem[Treml et~al.(2018)Treml, Gillman, Buskohl, and Vaia]{treml2018origami}
Benjamin Treml, Andrew Gillman, Philip Buskohl, and Richard Vaia.
\newblock Origami mechanologic.
\newblock \emph{Proceedings of the National Academy of Sciences}, 115\penalty0 (27):\penalty0 6916--6921, 2018.

\bibitem[Preston et~al.(2019)Preston, Rothemund, Jiang, Nemitz, Rawson, Suo, and Whitesides]{preston2019digital}
Daniel~J Preston, Philipp Rothemund, Haihui~Joy Jiang, Markus~P Nemitz, Jeff Rawson, Zhigang Suo, and George~M Whitesides.
\newblock Digital logic for soft devices.
\newblock \emph{Proceedings of the National Academy of Sciences}, 116\penalty0 (16):\penalty0 7750--7759, 2019.

\bibitem[Glozman et~al.(2010)Glozman, Hassidov, Senesh, and Shoham]{glozman2010self}
Daniel Glozman, Noam Hassidov, Merav Senesh, and Moshe Shoham.
\newblock A self-propelled inflatable earthworm-like endoscope actuated by single supply line.
\newblock \emph{IEEE Transactions on Biomedical Engineering}, 57\penalty0 (6):\penalty0 1264--1272, 2010.

\bibitem[Che et~al.(2018)Che, Yuan, Qi, and Meaud]{che2018viscoelastic}
Kaikai Che, Chao Yuan, H~Jerry Qi, and Julien Meaud.
\newblock Viscoelastic multistable architected materials with temperature-dependent snapping sequence.
\newblock \emph{Soft matter}, 14\penalty0 (13):\penalty0 2492--2499, 2018.

\bibitem[Milana et~al.(2022)Milana, Van~Raemdonck, Casla, De~Volder, Reynaerts, and Gorissen]{milana2022morphological}
Edoardo Milana, Bert Van~Raemdonck, Andrea~Serrano Casla, Michael De~Volder, Dominiek Reynaerts, and Benjamin Gorissen.
\newblock Morphological control of cilia-inspired asymmetric movements using nonlinear soft inflatable actuators.
\newblock \emph{Frontiers in Robotics and AI}, 8:\penalty0 788067, 2022.

\bibitem[Gude et~al.(2011)Gude, Hufenbach, and Kirvel]{gude2011piezoelectrically}
M~Gude, W~Hufenbach, and C~Kirvel.
\newblock Piezoelectrically driven morphing structures based on bistable unsymmetric laminates.
\newblock \emph{Composite structures}, 93\penalty0 (2):\penalty0 377--382, 2011.

\bibitem[Medina et~al.(2017)Medina, Gilat, and Krylov]{medina2017latching}
Lior Medina, Rivka Gilat, and Slava Krylov.
\newblock Latching in bistable electrostatically actuated curved micro beams.
\newblock \emph{International Journal of Engineering Science}, 110:\penalty0 15--34, 2017.

\bibitem[Hou et~al.(2018)Hou, Liu, Wan, Xu, Wen, Yu, Zhang, Li, and Chen]{hou2018magneto}
Xue Hou, Yin Liu, Guangchao Wan, Zhe Xu, Chunsheng Wen, Hui Yu, John~XJ Zhang, Jianbao Li, and Zi~Chen.
\newblock Magneto-sensitive bistable soft actuators: Experiments, simulations, and applications.
\newblock \emph{Applied Physics Letters}, 113\penalty0 (22):\penalty0 221902, 2018.

\bibitem[Harne et~al.(2013)Harne, Thota, and Wang]{harne2013concise}
RL~Harne, M~Thota, and KW~Wang.
\newblock Concise and high-fidelity predictive criteria for maximizing performance and robustness of bistable energy harvesters.
\newblock \emph{Applied Physics Letters}, 102\penalty0 (5):\penalty0 053903, 2013.

\bibitem[Betts et~al.(2012)Betts, Kim, Bowen, and Inman]{betts2012optimal}
David~N Betts, H~Alicia Kim, Christopher~R Bowen, and DJ~Inman.
\newblock Optimal configurations of bistable piezo-composites for energy harvesting.
\newblock \emph{Applied Physics Letters}, 100\penalty0 (11):\penalty0 114104, 2012.

\bibitem[Gorissen et~al.(2020)Gorissen, Melancon, Vasios, Torbati, and Bertoldi]{gorissen2020inflatable}
Benjamin Gorissen, David Melancon, Nikolaos Vasios, Mehdi Torbati, and Katia Bertoldi.
\newblock Inflatable soft jumper inspired by shell snapping.
\newblock \emph{Science Robotics}, 5\penalty0 (42):\penalty0 eabb1967, 2020.

\bibitem[Wang et~al.(2023)Wang, Zhang, Wang, Fang, Jiang, Yang, Zhu, Liu, Fan, and Kong]{wang2023untethered}
Zichao Wang, Xuan Zhang, Yang Wang, Ziyi Fang, He~Jiang, Qinglin Yang, Xuefeng Zhu, Mingze Liu, Xiaodong Fan, and Jie Kong.
\newblock Untethered soft microrobots with adaptive logic gates.
\newblock \emph{Advanced Science}, page 2206662, 2023.

\bibitem[Kaarthik et~al.(2022)Kaarthik, Sanchez, Avtges, and Truby]{kaarthik2022motorized}
Pranav Kaarthik, Francesco~L Sanchez, James Avtges, and Ryan~L Truby.
\newblock Motorized, untethered soft robots via 3d printed auxetics.
\newblock \emph{Soft Matter}, 18\penalty0 (43):\penalty0 8229--8237, 2022.

\bibitem[Zolfagharian et~al.(2020)Zolfagharian, Kaynak, and Kouzani]{zolfagharian2020closed}
Ali Zolfagharian, Akif Kaynak, and Abbas Kouzani.
\newblock Closed-loop 4d-printed soft robots.
\newblock \emph{Materials \& Design}, 188:\penalty0 108411, 2020.

\bibitem[Wang et~al.(2013)Wang, Chen, and Pham]{wang2013constant}
Dung-An Wang, Jyun-Hua Chen, and Huy-Tuan Pham.
\newblock A constant-force bistable micromechanism.
\newblock \emph{Sensors and Actuators A: Physical}, 189:\penalty0 481--487, 2013.

\bibitem[Shan et~al.(2015)Shan, Kang, Raney, Wang, Fang, Candido, Lewis, and Bertoldi]{shan2015multistable}
Sicong Shan, Sung~H Kang, Jordan~R Raney, Pai Wang, Lichen Fang, Francisco Candido, Jennifer~A Lewis, and Katia Bertoldi.
\newblock Multistable architected materials for trapping elastic strain energy.
\newblock \emph{Advanced Materials}, 27\penalty0 (29):\penalty0 4296--4301, 2015.

\bibitem[Ma et~al.(2019)Ma, Zhang, Zhang, Li, Wu, Jiang, and Chai]{ma2019origami}
Weili Ma, Zheng Zhang, Hao Zhang, Yang Li, Huaping Wu, Shaofei Jiang, and Guozhong Chai.
\newblock An origami-inspired cube pipe structure with bistable anti-symmetric cfrp shells driven by magnetic field.
\newblock \emph{Smart Materials and Structures}, 28\penalty0 (2):\penalty0 025028, 2019.

\bibitem[Wu et~al.(2018)Wu, Chaunsali, Yasuda, Yu, and Yang]{wu2018dial}
Ying Wu, Rajesh Chaunsali, Hiromi Yasuda, Kaiping Yu, and Jinkyu Yang.
\newblock Dial-in topological metamaterials based on bistable stewart platform.
\newblock \emph{Scientific reports}, 8\penalty0 (1):\penalty0 112, 2018.

\bibitem[Novelino et~al.(2020)Novelino, Ze, Wu, Paulino, and Zhao]{novelino2020untethered}
Larissa~S Novelino, Qiji Ze, Shuai Wu, Glaucio~H Paulino, and Ruike Zhao.
\newblock Untethered control of functional origami microrobots with distributed actuation.
\newblock \emph{Proceedings of the National Academy of Sciences}, 117\penalty0 (39):\penalty0 24096--24101, 2020.

\bibitem[Kaufmann et~al.(2022)Kaufmann, Bhovad, and Li]{kaufmann2022harnessing}
Joshua Kaufmann, Priyanka Bhovad, and Suyi Li.
\newblock Harnessing the multistability of kresling origami for reconfigurable articulation in soft robotic arms.
\newblock \emph{Soft Robotics}, 9\penalty0 (2):\penalty0 212--223, 2022.

\bibitem[Son et~al.(2022)Son, Park, Na, and Yoon]{son20224d}
Hyegyo Son, Yunha Park, Youngjin Na, and ChangKyu Yoon.
\newblock 4d multiscale origami soft robots: A review.
\newblock \emph{Polymers}, 14\penalty0 (19):\penalty0 4235, 2022.

\bibitem[Gillman et~al.(2018)Gillman, Wilson, Fuchi, Hartl, Pankonien, and Buskohl]{gillman2018design}
Andrew Gillman, Gregory Wilson, Kazuko Fuchi, Darren Hartl, Alexander Pankonien, and Philip Buskohl.
\newblock Design of soft origami mechanisms with targeted symmetries.
\newblock In \emph{Actuators}, volume~8, page~3. MDPI, 2018.

\bibitem[Yasuda et~al.(2017)Yasuda, Tachi, Lee, and Yang]{yasuda2017origami}
Hiromi Yasuda, Tomohiro Tachi, Mia Lee, and Jinkyu Yang.
\newblock Origami-based tunable truss structures for non-volatile mechanical memory operation.
\newblock \emph{Nature communications}, 8\penalty0 (1):\penalty0 962, 2017.

\bibitem[Gorissen et~al.(2019)Gorissen, Milana, Baeyens, Broeders, Christiaens, Collin, Reynaerts, and De~Volder]{gorissen2019hardware}
Benjamin Gorissen, Edoardo Milana, Arne Baeyens, Eva Broeders, Jeroen Christiaens, Klaas Collin, Dominiek Reynaerts, and Michael De~Volder.
\newblock Hardware sequencing of inflatable nonlinear actuators for autonomous soft robots.
\newblock \emph{Advanced Materials}, 31\penalty0 (3):\penalty0 1804598, 2019.

\bibitem[Melancon et~al.(2022)Melancon, Forte, Kamp, Gorissen, and Bertoldi]{melancon2022inflatable}
David Melancon, Antonio~Elia Forte, Leon~M Kamp, Benjamin Gorissen, and Katia Bertoldi.
\newblock Inflatable origami: Multimodal deformation via multistability.
\newblock \emph{Advanced Functional Materials}, page 2201891, 2022.

\bibitem[Hussein et~al.(2019)Hussein, Le~Moal, Younes, Bourbon, Haddab, and Lutz]{hussein2019design}
Hussein Hussein, Patrice Le~Moal, Rafic Younes, Gilles Bourbon, Yassine Haddab, and Philippe Lutz.
\newblock On the design of a preshaped curved beam bistable mechanism.
\newblock \emph{Mechanism and Machine Theory}, 131:\penalty0 204--217, 2019.

\bibitem[Salinas and Givli(2015)]{salinas2015can}
Gal Salinas and Sefi Givli.
\newblock Can a curved beam bistable mechanism have a secondary equilibrium that is more stable than its stress-free configuration?
\newblock \emph{Microsystem technologies}, 21:\penalty0 943--950, 2015.

\bibitem[Cadwell(1996)]{cadwell1996magnetic}
Louis~H Cadwell.
\newblock Magnetic damping: analysis of an eddy current brake using an airtrack.
\newblock \emph{American journal of physics}, 64\penalty0 (7):\penalty0 917--923, 1996.

\bibitem[Shmulevich et~al.(2014)Shmulevich, Joffe, Hotzen, and Elata]{shmulevich2014folded}
Shai Shmulevich, Aharon Joffe, Inbar Hotzen, and David Elata.
\newblock Are folded-beam suspensions really linear?
\newblock \emph{Procedia Engineering}, 87:\penalty0 624--627, 2014.

\bibitem[Becker et~al.(2003)Becker, Koehler, and Stone]{becker2003self}
Leif~E Becker, Stephan~A Koehler, and Howard~A Stone.
\newblock On self-propulsion of micro-machines at low reynolds number: Purcell's three-link swimmer.
\newblock \emph{Journal of fluid mechanics}, 490:\penalty0 15--35, 2003.

\bibitem[Kanso et~al.(2005)Kanso, Marsden, Rowley, and Melli-Huber]{kanso2005locomotion}
Eva Kanso, Jerrold~E Marsden, Clarence~W Rowley, and Juan~B Melli-Huber.
\newblock Locomotion of articulated bodies in a perfect fluid.
\newblock \emph{Journal of Nonlinear Science}, 15:\penalty0 255--289, 2005.

\bibitem[Gamus et~al.(2020)Gamus, Gat, and Or]{gamus2020dynamic}
Benny Gamus, Amir~D Gat, and Yizhar Or.
\newblock Dynamic inchworm crawling: Performance analysis and optimization of a three-link robot.
\newblock \emph{IEEE Robotics and Automation Letters}, 6\penalty0 (1):\penalty0 111--118, 2020.

\bibitem[Benichou and Givli(2013)]{BENICHOU201394}
Itamar Benichou and Sefi Givli.
\newblock Structures undergoing discrete phase transformation.
\newblock \emph{Journal of the Mechanics and Physics of Solids}, 61\penalty0 (1):\penalty0 94--113, 2013.
\newblock ISSN 0022-5096.
\newblock \doi{https://doi.org/10.1016/j.jmps.2012.08.009}.
\newblock URL \url{https://www.sciencedirect.com/science/article/pii/S0022509612001883}.

\bibitem[Nitecki and Givli(2021)]{NITECKI2021104634}
Saar Nitecki and Sefi Givli.
\newblock The mechanical behavior of 2-d lattices with bi-stable springs.
\newblock \emph{Journal of the Mechanics and Physics of Solids}, 157:\penalty0 104634, 2021.
\newblock ISSN 0022-5096.
\newblock \doi{https://doi.org/10.1016/j.jmps.2021.104634}.
\newblock URL \url{https://www.sciencedirect.com/science/article/pii/S0022509621002738}.

\bibitem[Puglisi and Truskinovsky(2000)]{puglisi2000mechanics}
G~Puglisi and Lev Truskinovsky.
\newblock Mechanics of a discrete chain with bi-stable elements.
\newblock \emph{Journal of the Mechanics and Physics of Solids}, 48\penalty0 (1):\penalty0 1--27, 2000.

\bibitem[Benichou et~al.(2013)Benichou, Faran, Shilo, and Givli]{benichou2013application}
Itamar Benichou, Eilon Faran, Doron Shilo, and Sefi Givli.
\newblock Application of a bi-stable chain model for the analysis of jerky twin boundary motion in nimnga.
\newblock \emph{Applied Physics Letters}, 102\penalty0 (1):\penalty0 011912, 2013.

\end{thebibliography}


\appendix


\end{document}